\RequirePackage[l2tabu, orthodox]{nag}
\documentclass[11pt]{article}

\usepackage[T1]{fontenc}

\usepackage[bitstream-charter]{mathdesign}
\usepackage{amsmath}
\usepackage[scaled=0.92]{PTSans}

\usepackage[
  paper  = letterpaper,
  left   = 1.65in,
  right  = 1.65in,
  top    = 1.0in,
  bottom = 1.0in,
  ]{geometry}

\usepackage[usenames,dvipsnames]{xcolor}
\definecolor{shadecolor}{gray}{0.9}

\usepackage[final,expansion=alltext]{microtype}
\usepackage[english]{babel}
\usepackage[parfill]{parskip}
\usepackage{afterpage}
\usepackage{framed}

\DeclareRobustCommand{\parhead}[1]{\textbf{#1}~}

\usepackage{lineno}

\usepackage{ragged2e}

\newcounter{parcount}

\usepackage{graphicx}
\usepackage[labelfont=bf]{caption}

\usepackage{booktabs}
\usepackage{multirow}

\usepackage{natbib}

\usepackage[algoruled]{algorithm2e}
\usepackage{listings}
\usepackage{fancyvrb}
\fvset{fontsize=\normalsize}

\usepackage[colorlinks,linktoc=all]{hyperref}
\usepackage[all]{hypcap}
\hypersetup{citecolor=MidnightBlue}
\hypersetup{linkcolor=black}
\hypersetup{urlcolor=MidnightBlue}

\usepackage[acronym,smallcaps,nowarn]{glossaries}

\lstdefinestyle{mystyle}{
    commentstyle=\color{OliveGreen},
    keywordstyle=\color{BurntOrange},
    numberstyle=\tiny\color{black!60},
    stringstyle=\color{MidnightBlue},
    basicstyle=\ttfamily,
    breakatwhitespace=false,
    breaklines=true,
    captionpos=b,
    keepspaces=true,
    numbers=left,
    numbersep=5pt,
    showspaces=false,
    showstringspaces=false,
    showtabs=false,
    tabsize=2
}
\usepackage[colorinlistoftodos,
           prependcaption,
           textsize=small,
           backgroundcolor=yellow,
           linecolor=lightgray,
           bordercolor=lightgray]{todonotes}

\DeclareMathOperator*{\argmax}{arg\,max}

\DeclareRobustCommand{\E}[2]{\mathbb{E}_{#1}\left[#2\right]}

\DeclareRobustCommand{\diag}[1]{\textrm{diag}\left(#1\right)}

\newcommand{\g}{\, | \,}

\newcommand{\Lcal}{\mathcal{L}}
\newcommand{\Ncal}{\mathcal{N}}

\newcommand{\bzero}{\mathbf{0}}

\newcommand{\bz}{\mathbf{z}}
\newcommand{\bx}{\mathbf{x}}
\newcommand{\bs}{\mathbf{s}}

\newcommand{\bI}{\mathbf{I}}

\newcommand{\beps}{\mathbf{\epsilon}}

\newcommand{\bmu}{\mathbf{\mu}}

\newcommand{\bepsilon}{\mathbf{\epsilon}}

\newcommand{\bsigma}{\bm{\sigma}}

\newcommand{\bSigma}{\bm{\Sigma}}

 \newacronym{ALI}{ali}{adversarially learned inference}
\newacronym{BIGAN}{bigan}{bidirectional generative adversarial network}
\newacronym{VI}{vi}{variational inference}
\newacronym{KL}{kl}{Kullback-Leibler}
\newacronym{ELBO}{elbo}{evidence lower bound}
\newacronym{MCMC}{mcmc}{Markov chain Monte Carlo}
\newacronym{HMC}{hmc}{Hamiltonian Monte Carlo}
\newacronym{RNN}{rnn}{recurrent neural network}
\newacronym{MLP}{mlp}{feed forward neural network}
\newacronym{VAE}{vae}{variational auto-encoder}
\newacronym{GAN}{gan}{generative adversarial network}
\newacronym{DCGAN}{dcgan}{deep convolutional generative adversarial network}
\newacronym{PresGAN}{presgan}{prescribed generative adversarial network}
\newacronym{DGM}{dgm}{deep generative model}
\newacronym{PGAN}{pgan}{prescribed generative adversarial network}
\newacronym{VEEGAN}{veegan}{veegan}
\newacronym{PACGAN}{pacgan}{packed {GAN}}
\newacronym{STYLEGAN}{stylegan}{Style {GAN}}
\newacronym{FID}{fid}{{F}r\'{e}chet {I}nception distance} 
\usepackage{subfigure}
\usepackage{authblk}
\usepackage{enumitem} 
\usepackage{bm}

\title{\textbf{Prescribed Generative Adversarial Networks}}

\author[1]{Adji B. Dieng}
\author[2, 3]{Francisco J. R. Ruiz}
\author[1, 2]{\\David M. Blei}
\author[4]{Michalis K. Titsias}
\affil[1]{Department of Statistics, Columbia University}
\affil[2]{Department of Computer Science, Columbia University}
\affil[3]{Department of Engineering, University of Cambridge}
\affil[4]{DeepMind}

\begin{document}
\maketitle

\begin{abstract}
\noindent \Glspl{GAN} are a powerful approach to unsupervised learning. They have achieved state-of-the-art performance in the image domain. However, \glspl{GAN} are limited in two ways. They often learn distributions with low support---a phenomenon known as mode collapse---and they do not guarantee the existence of a probability density, which makes evaluating generalization using predictive log-likelihood impossible. In this paper, we develop the prescribed \gls{GAN} (Pres\gls{GAN}) to address these shortcomings. Pres\glspl{GAN} add noise to the output of a density network and optimize an entropy-regularized adversarial loss. The added noise renders tractable approximations of the predictive log-likelihood and stabilizes the training procedure. The entropy regularizer encourages Pres\glspl{GAN} to capture all the modes of the data distribution. Fitting Pres\glspl{GAN} involves computing the intractable gradients of the entropy regularization term; Pres\glspl{GAN} sidestep this intractability using unbiased stochastic estimates. We evaluate Pres\glspl{GAN} on several datasets and found they mitigate mode collapse and generate samples with high perceptual quality. We further found that Pres\glspl{GAN} reduce the gap in performance in terms of predictive log-likelihood between traditional \glspl{GAN} and \glspl{VAE}.\footnote{\textbf{Code:} The code for this paper can be found at \url{https://github.com/adjidieng/PresGANs}.}\\

\end{abstract}

\section{Introduction}
\label{sec:introduction}
\glsresetall

\begin{figure*}[t]
	\centerline{\includegraphics[width=1.2\textwidth, height=4.2cm]{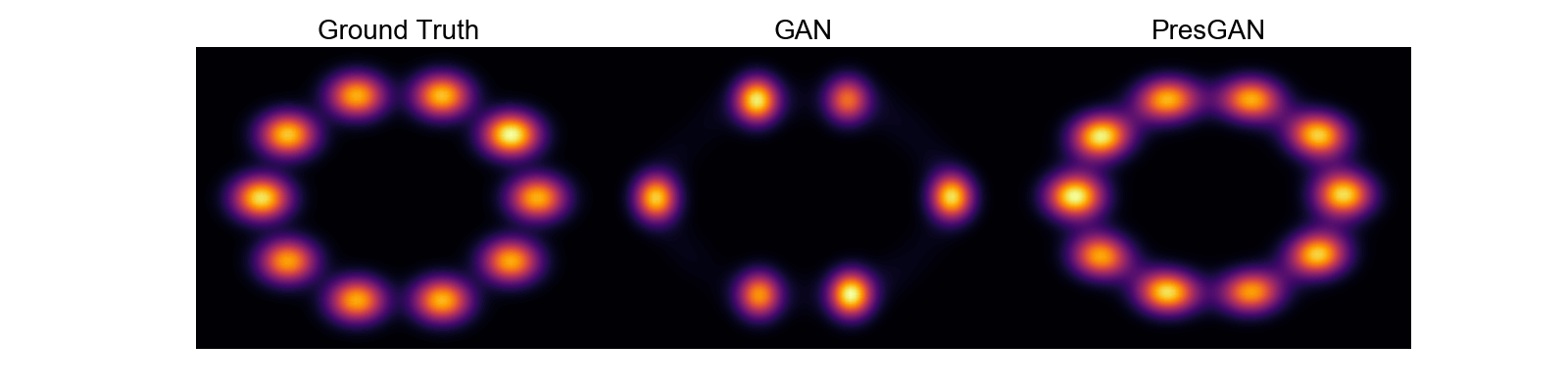}}
	\caption{Density estimation with \acrshort{GAN} and Pres\acrshort{GAN} on a toy two-dimensional experiment. The ground truth is a uniform mixture of $10$ Gaussians organized on a ring. Given the right set of hyperparameters, a \acrshort{GAN} could perfectly fit this target distribution. In this example we chose the \acrshort{GAN} hyperparameters such that it collapses---here $4$ out of $10$ modes are missing. We then fit the Pres\acrshort{GAN} using the same hyperparameters as the collapsing \acrshort{GAN}. The Pres\acrshort{GAN} is able to correct the collapsing behavior of the \acrshort{GAN} and learns a good fit for the target distribution.}
	\label{fig:toy_example_fig1}
\end{figure*}

\Glspl{GAN}~\citep{goodfellow2014generative} are a family of generative models that have shown great promise.
They achieve state-of-the-art performance in the image domain; for example image generation \citep{karras2019style, brock2018large}, image super-resolution \citep{ledig2017photo}, and image translation \citep{isola2017image}. 

\glspl{GAN} learn densities by defining a sampling procedure. A latent variable $\bz$ is sampled from a prior $p(\bz)$ and a sample $\tilde{\bx}(\bz; \theta)$ is generated by taking the output of a neural network with parameters $\theta$, called a generator, that takes $\bz$ as input. 
The density $p_{\theta}(\bx)$ implied by this sampling procedure is implicit and undefined \citep{mohamed2016learning}. However, \glspl{GAN} effectively learn the parameters $\theta$ by introducing a classifier $D_{\phi}$---a deep neural network with parameters $\phi$, called discriminator---that distinguishes between generated samples $\tilde{\bx}(\bz; \theta)$ and real data $\bx$, with distribution $p_d(\bx)$. The parameters $\theta$ and $\phi$ are learned jointly by optimizing the \gls{GAN} objective,
\begin{align}\label{eq:gan_loss}
		\Lcal_{\textrm{GAN}}(\theta, \phi) &= \; \E{\bx\sim p_d(\bx)}{\log D_{\phi}(\bx)} + \E{\bz\sim p(\bz)}{\log\left(1-D_{\phi}(\tilde{\bx}(\bz; \theta))\right)}.
\end{align}
\glspl{GAN} iteratively maximize the loss in Eq.\nobreakspace \ref {eq:gan_loss} with respect to $\phi$ and minimize it with respect to $\theta$. 

In practice, the minimax procedure described above is stopped when the generator produces realistic images. This is problematic because 
high perceptual quality does not necessarily correlate with goodness of fit to the target density. For example, memorizing the training data is a trivial solution to achieving high perceptual quality. Fortunately, \glspl{GAN} do not merely memorize the training data~\citep{zhang2017discrimination, arora2017generalization}. 

However \glspl{GAN} are able to produce images indistinguishable from real images while still failing to fully capture the target distribution~\citep{brock2018large, karras2019style}. Indeed \glspl{GAN} suffer from an issue known as \emph{mode collapse}. When mode collapse happens, the generative distribution $p_{\theta}(\bx)$ is degenerate and of low support~\citep{arora2017generalization, arora2018gans}. Mode collapse causes \glspl{GAN}, as density estimators, to fail both qualitatively and quantitatively. Qualitatively, mode collapse causes lack of diversity in the generated samples. This is problematic for certain applications of \glspl{GAN}, e.g. data augmentation. Quantitatively, mode collapse causes poor generalization to new data. This is because when mode collapse happens, there is a (support) mismatch between the learned distribution $p_{\theta}(\bx)$ and the data distribution. Using annealed importance sampling with a kernel density estimate of the likelihood, \citet{wu2016quantitative} report significantly worse log-likelihood scores for \glspl{GAN} when compared to \glspl{VAE}. Similarly poor generalization performance was reported by \citet{grover2018flow}. 

A natural way to prevent mode collapse in \glspl{GAN} is to maximize the entropy of the generator~\citep{belghazi2018mine}. Unfortunately the entropy of \glspl{GAN} is unavailable.
This is because the existence of the generative density $p_{\theta}(\bx)$ is not guaranteed \citep{mohamed2016learning, arjovsky2017wasserstein}.

In this paper, we propose a method to alleviate mode collapse in \glspl{GAN} resulting in a new family of \glspl{GAN} called prescribed \glspl{GAN} (Pres\glspl{GAN}). Pres\glspl{GAN} prevent mode collapse by explicitly maximizing the entropy of the generator. This is done by augmenting the loss in Eq.\nobreakspace \ref {eq:gan_loss} with the negative entropy of the generator, such that minimizing Eq.\nobreakspace \ref {eq:gan_loss} with respect to $\theta$ corresponds to fitting the data while also maximizing the entropy of the generative distribution. The existence of the generative density is guaranteed by adding noise to the output of a density network \citep{mackay1995bayesian, diggle1984monte}. This process defines the generative distribution $p_{\theta}(\bx)$, not as an implicit distribution as in standard \glspl{GAN}, but as an infinite mixture of well-defined densities as in continuous \glspl{VAE}~\citep{kingma2013auto, rezende2014stochastic}. The generative distribution of Pres\glspl{GAN} is therefore very flexible. 

Although the entropy of the generative distribution of Pres\glspl{GAN} is well-defined, it is intractable. However, fitting a Pres\gls{GAN} to data only involves computing the gradients of the entropy and not the entropy itself. Pres\glspl{GAN} use unbiased Monte Carlo estimates of these gradients. 

\parhead{An illustrative example.}
To demonstrate how Pres\glspl{GAN} alleviate mode collapse, we form a target distribution by organizing a uniform mixture of $K=10$ two-dimensional Gaussians on a ring. 
We draw $5{,}000$ samples from this target distribution. We first fit a \gls{GAN}, setting the hyperparameters so that the \gls{GAN} suffers from mode collapse\footnote{A \gls{GAN} can perfectly fit this distribution when choosing the right hyperparameters.}. We then use the same settings for Pres\gls{GAN} to assess whether it can correct the collapsing behavior of the \gls{GAN}. Figure\nobreakspace \ref {fig:toy_example_fig1} 
shows the collapsing behavior of the \gls{GAN}, which misses $4$ modes of the target distribution. The Pres\gls{GAN}, on the other hand, recovers all the modes. 
Section\nobreakspace \ref {sec:empirical} provides details about the settings of this synthetic experiment.

\parhead{Contributions.} This paper contributes to the literature on the two main open problems in the study of \glspl{GAN}: preventing mode collapse and evaluating log-likelihood.
\begin{itemize}[leftmargin=*]
	\item How can we perform entropy regularization of the generator of a \gls{GAN} so as to effectively prevent mode collapse? We achieve this by adding noise to the output of the generator; this ensures the existence of a density $p_{\theta}(\bx)$ and makes its entropy well-defined. We then regularize the \gls{GAN} loss to encourage densities $p_{\theta}(\bx)$ with high entropy. During training, we form unbiased estimators of the (intractable) gradients of the entropy regularizer. We show how this prevents mode collapse, as expected, in two sets of experiments (see Section\nobreakspace \ref {sec:empirical}). The first experiment follows the current standard for measuring mode collapse in the \gls{GAN} literature, which is to report the number of modes recovered by the \gls{GAN} on $\textsc{mnist}$ ($10$ modes) and $\textsc{stackedmnist}$ ($1{,}000$ modes) and the \gls{KL} divergence between the true label distribution and the one induced by the \gls{GAN}. We conducted a second experiment which sheds light on another way mode collapse can occur in \glspl{GAN}, which is when the data is imbalanced.
	\item How can we measure log-likelihood in \glspl{GAN}? Evaluating log-likelihood for \glspl{GAN} allows assessing how they generalize to new data. Existing measures focus on sample quality, which is not a measure of generalization. This inability to measure predictive log-likelihood for \glspl{GAN} has restricted their use to domains where one can use perceptual quality measures (e.g., the image domain). 
	Existing methods for evaluating log-likelihood for \glspl{GAN} either use a proxy to log-likelihood~\citep{sanchez2019out} or define the likelihood of the generator only at test time, which creates a mismatch between training and testing~\citep{wu2016quantitative}, or assume invertibility of the generator of the \gls{GAN}~\citep{grover2018flow}. Adding noise to the output of the generator immediately makes tractable predictive log-likelihood evaluation via importance sampling. 
\end{itemize} 

\parhead{Outline.} The rest of the paper is organized as follows. In Section\nobreakspace \ref {sec:background} we set the notation and provide desiderata for deep generative modeling. In Section\nobreakspace \ref {sec:method} we describe Pres\glspl{GAN} and how we compute their predictive log-likelihood to assess generalization. In Section\nobreakspace \ref {sec:related} we discuss related work. We then assess the performance of Pres\glspl{GAN} in terms of mode collapse, sample quality, and log-likelihood in Section\nobreakspace \ref {sec:empirical}. Finally, we conclude and discuss key findings in Section\nobreakspace \ref {sec:discussion}.
 
\section{Prologue}
\label{sec:background}

In this paper, we characterize a \gls{DGM} by its generative process and by the loss used to fit its parameters. We denote by $p_{\theta}(\bx)$ the generative distribution induced by the generative process---it is parameterized by a deep neural network with parameters $\theta$. The loss, that we denote by $\mathcal{L}(\theta, \phi)$, often requires an additional set of parameters $\phi$ that help learn the model parameters $\theta$. We next describe choices for $p_{\theta}(\bx)$ and $\mathcal{L}(\theta, \phi)$ and then specify desiderata for deep generative modeling.

\parhead{The generative distribution.} 
Recent \glspl{DGM} define the generative distribution either as an implicit distribution or as an infinite mixture~\citep{goodfellow2014generative, kingma2013auto, rezende2014stochastic}. 

Implicit generative models define a density using a sampling procedure. This is the approach of \glspl{GAN}~\citep{goodfellow2014generative}. A latent variable $\bz$ is sampled from a prior $p(\bz)$, usually a standard Gaussian or a uniform distributon, and a sample is generated by taking the output of a neural network that takes $\bz$ as input. The density $p_{\theta}(\bx)$ implied by this sampling procedure is undefined. Any measure that relies on an analytic form of the density $p_{\theta}(\bx)$ is therefore unavailable; e.g., the log-likelihood or the entropy.

An alternative way to define the generative distribution is by using the approach of \glspl{VAE}~\citep{kingma2013auto, rezende2014stochastic}. They define $p_{\theta}(\bx)$ as an infinite mixture,
\begin{align}\label{eq:vaes}
	p_{\theta}(\bx) = \int_{}^{} p_{\theta}(\bx \g \bz) \; p(\bz)\; d\bz.
\end{align}
Here the mixing distribution is the prior $p(\bz)$. The conditional distribution $p_{\theta}(\bx \g \bz)$ is an exponential family distribution, such as a Gaussian or a Bernoulli, parameterized by a neural network with parameters $\theta$. Although both the prior $p(\bz)$ and $p_{\theta}(\bx \g \bz)$ are simple tractable distributions, the generative distribution $p_{\theta}(\bx)$ is highly flexible albeit intractable. Because $p_{\theta}(\bx)$ in Eq.\nobreakspace \ref {eq:vaes} is well-defined, the log-likelihood and the entropy are also well-defined (although they may be analytically intractable). 

\parhead{The loss function.} 
Fitting the models defined above requires defining a learning procedure by specifying a loss function. \glspl{GAN} introduce a classifier $D_{\phi}$, a deep neural network parameterized by $\phi$, to discriminate between samples from the data distribution $p_d(\bx)$ and the generative distribution $p_{\theta}(\bx)$. The auxiliary parameters $\phi$ are learned jointly with the model parameters $\theta$ by optimizing the loss in Eq.\nobreakspace \ref {eq:gan_loss}. This training procedure leads to high sample quality but often suffers from \emph{mode collapse} \citep{arora2017generalization, arora2018gans}. 

An alternative approach to learning $\theta$ is via maximum likelihood. This requires a well-defined density $p_{\theta}(\bx)$ such as the one in Eq.\nobreakspace \ref {eq:vaes}. Although well-defined, $p_{\theta}(\bx)$ is intractable, making it difficult to learn the parameters $\theta$ by maximum likelihood. \glspl{VAE} instead introduce a recognition network---a neural network with parameters $\phi$ that takes data $\bx$ as input and outputs a distribution over the latent variables $\bz$---and maximize a lower bound on $\log p_{\theta}(\bx)$ with respect to both $\theta$ and $\phi$,
\begin{align}\label{eq:elbo}
	\Lcal_{\textrm{VAE}}(\theta, \phi) &= E_{p_d(\bx)} E_{q_{\phi}(\bz \g \bx)}\left[ \log \frac{p_{\theta}(\bx , \bz)}{q_{\phi}(\bz \g \bx)} \right] = -\text{KL}(q_{\phi}(\bz \g \bx) p_d(\bx) \vert\vert p_{\theta}(\bx , \bz)).
\end{align}
Here $\text{KL}(\cdot \vert\vert \cdot)$ denotes the \gls{KL} divergence. Maximizing $\Lcal_{\textrm{VAE}}(\theta, \phi)$ is equivalent to minimizing this \gls{KL} which leads to issues such as latent variable collapse~\citep{bowman2015generating, dieng2018avoiding}. Furthermore, optimizing Eq.\nobreakspace \ref {eq:elbo} may lead to blurriness in the generated samples because of a property of the reverse $\gls{KL}$ known as \emph{zero-forcing} \citep{minka2005divergence}.

\parhead{Desiderata.} 
We now outline three desiderata for \glspl{DGM}.

\emph{High sample quality}. A \gls{DGM} whose parameters $\theta$ have been fitted using real data should generate new data with the same qualitative precision as the data it was trained with. For example, if a \gls{DGM} is trained on a dataset composed of human faces, it should generate data with all features that make up a face at the same resolution as the training data.

\emph{High sample diversity}. High sample quality alone is not enough. For example, a degenerate \gls{DGM} that is only able to produce one single sample is not desirable, even if the sample quality is perfect.
Therefore we require sample diversity; a \gls{DGM} should ideally capture all modes of the data distribution.

\emph{Tractable predictive log-likelihood}. \glspl{DGM} are density estimators and as such we should evaluate how they generalize to new data. High sample quality and diversity are not measures of generalization. We therefore require tractable predictive log-likelihood as a desideratum for deep generative modeling.

We next introduce a new family of \glspl{GAN} that fulfills all the desiderata. 
 
\section{Prescribed Generative Adversarial Networks}
\label{sec:method}

Pres\glspl{GAN} generate data following the generative distribution in Eq.\nobreakspace \ref {eq:vaes}. 
Note that this generative process is the same as for standard \glspl{VAE} \citep{kingma2013auto,rezende2014stochastic}.
In particular, Pres\glspl{GAN} set the prior $p(\bz)$ and the likelihood $p_{\theta}(\bx\g \bz)$ to be Gaussians, 
\begin{equation}\label{eq:semi_implicit}
	p(\bz) = \mathcal{N}(\bz\g\bzero, \bI) \quad \text{and} \quad p_{\theta}(\bx\g \bz) = \Ncal\left(\bx\g \bmu_{\theta}(\bz), \bSigma_{\theta}(\bz) \right).
\end{equation}
The mean $\bmu_{\theta}(\bz)$ and covariance $\bSigma_{\theta}(\bz)$ of the conditional $p_{\theta}(\bx\g \bz)$ are given by a neural network that takes $\bz$ as input.

In general, both the mean $\bmu_{\theta}(\bz)$ and the covariance $\bSigma_{\theta}(\bz)$ can be functions of $\bz$. For simplicity, in order to speed up the learning procedure, we set
the covariance matrix to be diagonal with elements independent from $\bz$, i.e., $\bSigma_{\theta}(\bz)=\diag{\bsigma^2}$, and we learn the vector $\bsigma$ together with $\theta$. From now on, we parameterize the mean with $\eta$, write $\bmu_{\eta}(\bz)$, and define $\theta = (\eta, \bsigma)$ as the parameters of the generative distribution.

To fit the model parameters $\theta$, Pres\glspl{GAN} optimize an adversarial loss similarly to \glspl{GAN}.
In doing so, they keep \glspl{GAN}' ability to generate samples with high perceptual quality. Unlike \glspl{GAN}, the entropy of the generative distribution of Pres\glspl{GAN} is well-defined, and therefore Pres\glspl{GAN} can prevent mode collapse by adding an entropy regularizer to Eq.\nobreakspace \ref {eq:gan_loss}. Furthermore, because Pres\glspl{GAN} define a density over their generated samples, we can measure how they generalize to new data using predictive log-likelihood. 
We describe the entropy regularization in Section\nobreakspace \ref {subsec:entropy_reg} and how to approximate the predictive log-likelihood in Section\nobreakspace \ref {subsec:loglik_evaluation}.

\subsection{Avoiding mode collapse via entropy regularization}
\label{subsec:entropy_reg}

One of the major issues that \glspl{GAN} face is mode collapse, where the generator tends to model only some parts or modes of the data distribution \citep{arora2017generalization, arora2018gans}. 
Pres\glspl{GAN} mitigate this problem by explicitly maximizing the entropy of the generative distribution,
\begin{equation}\label{eq:sigan_loss}
	\Lcal_{\text{Pres}\textrm{GAN}}(\theta, \phi) = \Lcal_{\textrm{GAN}}(\theta, \phi) - \lambda \mathcal{H}\left(p_{\theta}(\bx)\right).
\end{equation}
Here $\mathcal{H}\left(p_{\theta}(\bx)\right)$ denotes the entropy of the generative distribution. It is defined as
\begin{equation}\label{eq:def_entropy}
	\mathcal{H}\left(p_{\theta}(\bx)\right) = -\E{p_{\theta}(\bx)}{\log p_{\theta}(\bx)}.
\end{equation}
The loss $\Lcal_{\textrm{GAN}}(\theta, \phi)$ in Eq.\nobreakspace \ref {eq:sigan_loss} can be that of any of the existing \gls{GAN} variants. In Section\nobreakspace \ref {sec:empirical} we explore the standard \gls{DCGAN}~\citep{radford2015unsupervised} and the more recent Style\gls{GAN}~\citep{karras2019style} architectures. 

The constant $\lambda$ in Eq.\nobreakspace \ref {eq:sigan_loss} is a hyperparameter that controls the strength of the entropy regularization. 
In the extreme case when $\lambda=0$, the loss function of Pres\gls{GAN} coincides with the loss of a \gls{GAN}, where we replaced its implicit generative distribution with the infinite mixture in Eq.\nobreakspace \ref {eq:vaes}. In the other extreme when $\lambda=\infty$, optimizing $\Lcal_{\text{Pres}\textrm{GAN}}(\theta, \phi)$ corresponds to fitting a maximum entropy generator that ignores the data. For any intermediate values of $\lambda$, the first term of $\Lcal_{\text{Pres}\textrm{GAN}}(\theta, \phi)$ encourages the generator to fit the data distribution, whereas the second term encourages to cover all of the modes of the data distribution.

The entropy $\mathcal{H}\left(p_{\theta}(\bx)\right)$ is intractable because the integral in Eq.\nobreakspace \ref {eq:def_entropy} cannot be computed. However, fitting the parameters $\theta$ of Pres\glspl{GAN} only requires the gradients of the entropy. In Section\nobreakspace \ref {sec:fitting} we describe how to form unbiased Monte Carlo estimates of these gradients.

\subsection{Fitting Prescribed Generative Adversarial Networks}\label{sec:fitting}
We fit Pres\glspl{GAN} following the same adversarial procedure used in \glspl{GAN}. That is, we alternate between updating the parameters of the generative distribution $\theta$ and the parameters of the discriminator $\phi$. The full procedure is given in Algorithm\nobreakspace \ref {alg:full_algo}.
We now describe each part in detail. 

\begin{algorithm}[t]
  \SetAlgoNoLine
    \DontPrintSemicolon
    \SetKwInOut{KwInput}{input}
    \SetKwInOut{KwOutput}{output}
    \KwInput{Data $\bx$, entropy regularization level $\lambda$}
    Initialize parameters $\eta, \bsigma, \phi$\;
    \For{\emph{iteration} $t=1,2,\ldots$}{
      Draw minibatch of observations $\bx_1, \dots, \bx_b, \dots, \bx_B$\;
       \For{$b=1,2,\ldots, B$}{
      	Get noised data: $ \bepsilon_b \sim \mathcal{N}(\bzero, \bI)$ and $\widehat{\bx}_b = \bx_b + \bsigma \odot \bepsilon_b$\;
      	Draw latent variable $\bz_b \sim \mathcal{N}(\bzero, \bI)$\;
      	Generate data: $ \bs_b \sim \mathcal{N}(\bzero, \bI)$ and $\tilde{\bx}_b=\tilde{\bx}_b(\bz_b,\bs_b; \theta) = \bmu_{\eta}(\bz_b) + \bsigma \odot \bs_b$\;
	}
      Compute $\nabla_{\phi}\Lcal_{\text{Pres}\textrm{GAN}}(\theta, \phi)$ (Eq.\nobreakspace \ref {eq:grad_phi}) and take a gradient step for $\phi$\;
      Initialize an \acrshort{HMC} sampler using $\bz_b$\;
      Draw $\tilde{\bz}_b^{(m)} \sim p_{\theta}(\bz\g \tilde{\bx}_b)$ for $m = 1, \dots, M$ and $b = 1, \dots, B$ using that sampler\;
      Compute $\widehat{\nabla}_{\eta}\Lcal_{\text{Pres}\textrm{GAN}}((\eta, \bsigma), \phi)$ (Eq.\nobreakspace \ref {eq:grad_eta}) and 
      take a gradient step for $\eta$\;
      Compute $\widehat{\nabla}_{\bsigma}\Lcal_{\text{Pres}\textrm{GAN}}((\eta, \bsigma), \phi)$ (Eq.\nobreakspace \ref {eq:grad_sigma}) and 
      take a gradient step for $\bsigma$\;
      Truncate $\bsigma$ in the range $[\bsigma_{\text{low}}, \bsigma_{\text{high}}]$\;
    }
    \caption{Learning with Prescribed Generative Adversarial Networks (Pres\glspl{GAN}) \label{alg:full_algo}}
 \end{algorithm}

\parhead{Fitting the generator.}
We fit the generator using stochastic gradient descent. This requires computing the gradients of the Pres\gls{GAN} loss with respect to $\theta$,
\begin{equation}\label{eq:grad_theta}
	\nabla_{\theta}\Lcal_{\text{Pres}\textrm{GAN}}(\theta, \phi) = \nabla_{\theta}\Lcal_{\textrm{GAN}}(\theta, \phi) - \lambda \nabla_{\theta} \mathcal{H}\left(p_{\theta}(\bx)\right).
\end{equation}
We form stochastic estimates of $\nabla_{\theta}\Lcal_{\textrm{GAN}}(\theta, \phi)$ based on reparameterization \citep{kingma2013auto,rezende2014stochastic,titsias2014doubly}; this requires differentiating Eq.\nobreakspace \ref {eq:gan_loss}. Specifically, we introduce a noise variable $\bepsilon$ to reparameterize the conditional from Eq.\nobreakspace \ref {eq:semi_implicit},\footnote{With this reparameterization we use the notation $\bx(\bz, \beps; \theta)$ instead of $\tilde{\bx}(\bz; \theta)$ to denote a sample from the generative distribution.}
\begin{align}\label{eq:sigan_sample}
	\bx(\bz, \beps; \theta) &= \bmu_{\eta}(\bz) + \bsigma \odot \bepsilon,
\end{align}
where $\theta = (\eta, \bsigma)$ and $\bepsilon \sim \mathcal{N}(\bzero, \bI)$. Here $\bmu_{\eta}(\bz)$ and $\bsigma$ denote the mean and standard deviation of the conditional $p_{\theta}(\bx\g\bz)$, respectively. 
We now write the first term of Eq.\nobreakspace \ref {eq:grad_theta} as an expectation with respect to the latent variable $\bz$ and the noise variable $\beps$ and push the gradient into the expectation,
\begin{align}\label{eq:grad_gan}
	\nabla_{\theta}\Lcal_{\textrm{GAN}}(\theta, \phi) &= \mathbb{E}_{p(\bz) p(\bepsilon)} \left[\nabla_{\theta} \log\left(1-D_{\phi}(\bx(\bz, \beps; \theta))\right)\right]
	.
\end{align}
In practice we use an estimate of Eq.\nobreakspace \ref {eq:grad_gan} using one sample from $p(\bz)$ and one sample from $p(\bepsilon)$,
\begin{align}\label{eq:estimate_grad_gan}
		\widehat{\nabla}_{\theta} \Lcal_{\textrm{GAN}}(\theta, \phi) &= \nabla_{\theta} \log\left(1-D_{\phi}(\bx(\bz, \beps; \theta))\right)
		.
\end{align}
The second term in Eq.\nobreakspace \ref {eq:grad_theta}, corresponding to the gradient of the entropy, is intractable.  
We estimate it using the same approach as \citet{titsias2018unbiased}. 
We first use the reparameterization in Eq.\nobreakspace \ref {eq:sigan_sample} to express the gradient of the entropy as an expectation,
\begin{align*}
    \nabla_{\theta} \mathcal{H}\left(p_{\theta}(\bx)\right) 
    &= - \nabla_{\theta} \E{p_{\theta}(\bx)}{\log p_{\theta}(\bx)}
    = - \nabla_{\theta} \E{p(\bepsilon)p(\bz)}{\log p_{\theta}(\bx)\big|_{\bx = \bx(\bz, \beps; \theta)}} \\
    & = - \E{p(\bepsilon)p(\bz)}{\nabla_{\theta} \log p_{\theta}(\bx)\big|_{\bx=\bx(\bz, \beps; \theta)}}\\
    &= - \E{p(\bepsilon)p(\bz)}{\nabla_{\bx} \log p_{\theta}(\bx)\big|_{\bx=\bx(\bz, \beps; \theta)} \nabla_{\theta}\bx(\bz,\bepsilon; \theta)},
\end{align*}
where we have used the score function identity $\E{p_{\theta}(\bx)}{\nabla_{\theta}\log p_{\theta}(\bx)}=0$ on the second line.
We form a one-sample estimator of the gradient of the entropy as\looseness=-1
\begin{equation}\label{eq:estimate_grad_entropy}
		\widehat{\nabla}_{\theta} \mathcal{H}\left(p_{\theta}(\bx)\right) 
		=  -\nabla_{\bx}\! \log p_{\theta}(\bx)\big|_{\bx = \bx(\bz,\bepsilon; \theta)} \times \nabla_{\theta} \bx(\bz,\bepsilon; \theta).
\end{equation}
In Eq.\nobreakspace \ref {eq:estimate_grad_entropy}, the gradient with respect to the reparameterization transformation $\nabla_{\theta} \bx(\bz,\bepsilon; \theta)$ is tractable and can be obtained via back-propagation.
We now derive $\nabla_{\bx} \log p_{\theta}(\bx)$,
\begin{align*}
	\nabla_{\bx} \log p_{\theta}(\bx) & = \frac{\nabla_{\bx}p_{\theta}(\bx)}{p_{\theta}(\bx)} = \frac{\int_{}^{}\nabla_{\bx}p_{\theta}(\bx, \bz) d\bz}{p_{\theta}(\bx)}
	= \int_{}^{}\frac{\frac{\nabla_{\bx}p_{\theta}(\bx \g \bz)}{p_{\theta}(\bx \g \bz)} p_{\theta}(\bx , \bz)}{p_{\theta}(\bx)}d\bz \\
	&= \int_{}^{} \nabla_{\bx}\log p_{\theta}(\bx \g \bz) p_{\theta}(\bz \g \bx) d\bz = \E{p_{\theta}(\bz \g \bx)}{\nabla_{\bx}\log p_{\theta}(\bx \g \bz)}.
\end{align*}
While this expression is still intractable, we can estimate it. One way is to use self-normalized importance sampling with a proposal learned using moment matching with an encoder~\citep{dieng2019reweighted}. However, this would lead to a biased (albeit asymptotically unbiased) estimate of the entropy. In this paper, we form an unbiased estimate of $\nabla_{\bx} \log p_{\theta}(\bx)$ using samples $\bz^{(1)}, \dots, \bz^{(M)}$ from the posterior,
\begin{equation}\label{eq:estimate_logdensity}
	\widehat{\nabla}_{\bx} \log p_{\theta}(\bx) = \frac{1}{M}\sum_{m=1}^{M}\nabla_{\bx} \log p_{\theta}(\bx \g \bz^{(m)}),
  \qquad
  \bz^{(m)}\sim p_{\theta}(\bz \g \bx). 
\end{equation}
We obtain these samples using \gls{HMC} \citep{neal2011mcmc}.
Crucially, in order to speed up the algorithm, we initialize the \gls{HMC} sampler at stationarity. That is, we initialize the \gls{HMC} sampler with the sample $\bz$ that was used to produce the generated sample $\bx(\bz,\bepsilon; \theta)$ in Eq.\nobreakspace \ref {eq:sigan_sample}, which by construction is an exact sample from $p_{\theta}(\bz \g \bx)$. 
This implies that only a few \gls{HMC} iterations suffice to get good estimates of the gradient \citep{titsias2018unbiased}. We also found this holds empirically; for example in Section\nobreakspace \ref {sec:empirical} we use $2$ burn-in iterations and $M=2$ \gls{HMC} samples to form the Monte Carlo estimate in Eq.\nobreakspace \ref {eq:estimate_logdensity}.

Finally, using 
Eqs.\nobreakspace \ref {eq:grad_theta} and\nobreakspace   \ref {eq:estimate_grad_gan} to\nobreakspace  \ref {eq:estimate_logdensity} 
we can approximate the gradient of the entropy-regularized adversarial loss with respect to the model parameters $\theta$, 
\begin{align}\label{eq:grad_theta2}
  \widehat{\nabla}_{\theta}\Lcal_{\text{Pres}\textrm{GAN}}(\theta, \phi) &= \nabla_{\theta} \log\left(1-D_{\phi}(\bx(\bz, \beps; \theta))\right) \nonumber\\
  &+ \frac{\lambda}{M}\sum_{m=1}^{M} \nabla_{\bx} \log p_{\theta}(\bx \g \bz^{(m)})\big|_{\bx = \bx(\bz^{(m)},\bepsilon;\theta)} \times \nabla_{\theta} \bx\left(\bz^{(m)},\bepsilon;\theta\right).
\end{align}
In particular, the gradient with respect to the generator's parameters $\eta$ is unbiasedly approximated by
\begin{align}\label{eq:grad_eta}
  \widehat{\nabla}_{\eta}\Lcal_{\text{Pres}\textrm{GAN}}(\theta, \phi) &= \nabla_{\eta} \log\left(1-D_{\phi}(\bx(\bz, \beps; \theta))\right) \nonumber\\
  &- \frac{\lambda}{M}\sum_{m=1}^{M} \frac{\bx(\bz^{(m)},\bepsilon;\theta) - \bmu_{\eta}\left(\bz^{(m)}\right)}{\bsigma^2} \nabla_{\eta} \bmu_{\eta}(\bz^{(m)}),
\end{align}
and the gradient estimator with respect to the standard deviation $\bsigma$ is 
\begin{align}\label{eq:grad_sigma}
  \widehat{\nabla}_{\bsigma}\Lcal_{\text{Pres}\textrm{GAN}}(\theta, \phi) &= \nabla_{\bsigma} \log\left(1-D_{\phi}(\bx(\bz, \beps; \theta))\right) \nonumber\\
  &- \frac{\lambda}{M}\sum_{m=1}^{M} \frac{\bx(\bz^{(m)},\bepsilon;\theta) - \bmu_{\eta}\left(\bz^{(m)}\right)}{\bsigma^2} \cdot \bepsilon.
\end{align}
These gradients are used in a stochastic optimization algorithm to fit the generative distribution of Pres\gls{GAN}.

\parhead{Fitting the discriminator.}
Since the entropy term in Eq.\nobreakspace \ref {eq:sigan_loss} does not depend on $\phi$, optimizing the discriminator of a Pres\gls{GAN} is analogous to optimizing the discriminator of a \gls{GAN},
\begin{align}\label{eq:grad_phi}
\nabla_{\phi}\Lcal_{\text{Pres}\textrm{GAN}}(\theta, \phi) = \nabla_{\phi}\Lcal_{\textrm{GAN}}(\theta, \phi).
\end{align}

To prevent the discriminator from getting stuck in a bad local optimum where it can perfectly distinguish between real and generated data by relying on the added noise, we apply the same amount of noise to the real data $\bx$ as the noise added to the generated data. That is, when we train the discriminator we corrupt the real data according to
\begin{align}\label{eq:noised_data}
	\widehat{\bx} &= \bx + \bsigma \odot \bepsilon,
\end{align}
where $\bsigma$ is the standard deviation of the generative distribution and $\bx$ denotes the real data. 
We then let the discriminator distinguish between $\widehat{\bx}$ and $\bx(\bz,\bepsilon; \theta)$ from Eq.\nobreakspace \ref {eq:sigan_sample}.

This data noising procedure is a form of \emph{instance noise} \citep{sonderby2016amortised}. However, instead of using a fixed annealing schedule for the noise variance as \citet{sonderby2016amortised}, we let $\bsigma$ be part of the parameters of the generative distribution and fit it using gradient descent according to Eq.\nobreakspace \ref {eq:grad_sigma}.

\parhead{Stability.} Data noising stabilizes the training procedure and prevents the discriminator from perfectly being able to distinguish between real and generated samples using the background noise. We refer the reader to \citet{ferenc2016instance} for a detailed exposition. 

When fitting Pres\glspl{GAN}, data noising is not enough to stabilize training. This is because there are two failure cases brought in by learning the variance $\bsigma^2$ using gradient descent. The first failure mode is when the variance gets very large, leading to a generator completely able to fool the discriminator. Because of data noising, the discriminator cannot distinguish between real and generated samples when the variance of the noise is large. 

The second failure mode is when $\bsigma^2$ gets very small, which makes the gradient of the entropy in Eq.\nobreakspace \ref {eq:grad_eta} dominate the overall gradient of the generator. This is problematic because the learning signal from the discriminator is lost. 

To stabilize training and avoid the two failure cases discussed above we truncate the variance of the generative distribution,
$\bsigma_{\text{low}} \leq \bsigma \leq \bsigma_{\text{high}}$ (we apply this truncation element-wise). The limits $\bsigma_{\text{low}}$ and $\bsigma_{\text{high}}$ are hyperparameters.

\subsection{Enabling tractable predictive log-likelihood approximation}
\label{subsec:loglik_evaluation}

Replacing the implicit generative distribution of \glspl{GAN} with the infinite mixture distribution defined in Eq.\nobreakspace \ref {eq:vaes} has the advantage that the predictive log-likelihood can be tractably approximated. 
Consider an unseen datapoint $\bx^*$. We estimate its log marginal likelihood $\log p_{\theta}(\bx^*)$ using importance sampling,
\begin{equation}\label{eq:loglik}
  \log p_{\theta}(\bx^*) \approx \log \left( \frac{1}{S} \sum_{s=1}^{S} \frac{p_{\theta}\left(\bx^*\g \bz^{(s)}\right)\cdot p\left(\bz^{(s)}\right)}{r\left(\bz^{(s)}\g \bx^*\right)}\right),
\end{equation}
where we draw $S$ samples $\bz^{(1)}, \dots, \bz^{(S)}$ from a proposal distribution $r(\bz\g \bx^*)$.

There are different ways to form a good proposal $r(\bz\g \bx^*)$, and we discuss several alternatives in Section\nobreakspace \ref {app:proposals} of the appendix. In this paper, we take the following approach. We define the proposal as a Gaussian distribution,
\begin{align}
  r(\bz\g \bx^*) &= \mathcal{N}(\bmu_r, \bSigma_r) 
  .
\end{align}
We set the mean parameter $\bmu_r$ to the \emph{maximum a posteriori} solution, i.e., $\bmu_r = \argmax_z \left( \log p_{\theta}\left(\bx^*\g \bz \right) + \log p\left(\bz\right) \right)$. We initialize this maximization algorithm using the mean of a pre-fitted encoder, $q_{\gamma}(\bz\g \bx^*)$. The encoder is fitted by minimizing the reverse \gls{KL} divergence between $q_{\gamma}(\bz\g \bx)$ and the true posterior $p_{\theta}(\bz\g \bx)$ using the training data. This \gls{KL} is 
\begin{align}\label{eq:kl}
  &\gls{KL}\left( q_{\gamma}(\bz\g \bx) \vert\vert p_{\theta}(\bz\g \bx) \right) 
  = \log p_{\theta}(\bx) - \mathbb{E}_{q_{\gamma}(\bz\g \bx)}\left[ \log p_{\theta}(\bx \g \bz)p(\bz) - \log q_{\gamma}(\bz\g \bx) \right]
  .
\end{align}
Because the generative distribution is fixed at test time, minimizing the \gls{KL} here is equivalent to maximizing the second term in Eq.\nobreakspace \ref {eq:kl}, which is the \gls{ELBO} objective of \glspl{VAE}.

We set the proposal covariance $\bSigma_r$ as an overdispersed version\footnote{In general, overdispersed proposals lead to better importance sampling estimates.} of the encoder's covariance matrix, which is diagonal. In particular, to obtain $\bSigma_r$ we multiply the elements of the encoder's covariance by a factor $\gamma$. In Section\nobreakspace \ref {sec:empirical} we set $\gamma$ to $1.2$.
 
\section{Related Work}
\label{sec:related}

\glspl{GAN} \citep{goodfellow2014generative} have been extended in multiple ways, using alternative distance metrics and optimization methods \citep[see, e.g.,][]{li2015generative,dziugaite2015training,nowozin2016f,arjovsky2017wasserstein,ravuri2018learning,genevay2017learning} or using ideas from \glspl{VAE}~\citep{makhzani2015adversarial, mescheder2017adversarial, dumoulin2016adversarially, donahue2016adversarial, tolstikhin2017wasserstein, ulyanov2018takes, rosca2017variational}.

Other extensions aim at improving the sample diversity of \glspl{GAN}. For example, \citet{srivastava2017veegan} 
use a reconstructor network that reverses the action of the generator. \citet{lin2018pacgan} use multiple observations (either real or generated) as an input to the discriminator to prevent mode collapse. \citet{azadi2018discriminator} and \citet{turner2018metropolis} use sampling mechanisms to correct errors of the generative distribution. \citet{xiao2018bourgan} relies on identifying the geometric structure of the data embodied under a specific distance metric.
Other works have combined adversarial learning with maximum likelihood \citep{grover2018flow, yin2019semi}; however, the low sample quality induced by maximum likelihood still occurs. Finally, \citet{cao2018improving} introduce a regularizer for the discriminator to encourage diverse activation patterns in the discriminator across different samples. 
In contrast to these works, Pres\glspl{GAN} regularize the entropy of the generator to prevent mode collapse. 

The idea of entropy regularization has been widely applied in many problems that involve estimation of unknown probability distributions. Examples include approximate Bayesian inference, where the variational objective contains an entropy penalty \citep{jordan1998learning,bishop2006pattern,wainwright2008graphical,blei2017variational}; reinforcement learning, where the entropy regularization allows to estimate more uncertain and explorative policies \citep{schulman2015trust, mnih2016asynchronous}; statistical learning, where entropy regularization allows an inferred probability distribution to  avoid collapsing to a deterministic solution \citep{freund1997decision, soofi2000principal, jaynes2003probability}; or optimal transport \citep{rigollet2018entropic}. More recently, \citet{kumar2019maximum} have developed maximum-entropy generators for energy-based models using mutual information as a proxy for entropy. 

Another body of related work is about how to quantitatively evaluate \glspl{GAN}. Inception scores measure the sample quality of \glspl{GAN} and are used extensively in the \gls{GAN} literature~\citep{salimans2016improved, heusel2017gans, binkowski2018demystifying}. However, sample quality measures only assess the quality of \glspl{GAN} as data generators and not as density estimators. Density estimators are evaluated for generalization to new data. Predictive log-likelihood is a measure of goodness of fit that has been used to assess generalization; for example in \glspl{VAE}. Finding ways to evaluate predictive log-likelihood for \glspl{GAN} has been an open problem, because \glspl{GAN} do not define a density on the generated samples. 
\citet{wu2016quantitative} use a kernel density estimate~\citep{parzen1962estimation} and estimate the log-likelihood with annealed importance sampling~\citep{neal2001annealed}.
\citet{balaji2018entropic} show that an optimal transport \gls{GAN} with entropy regularization can be viewed as a generative model that maximizes a variational lower bound on average sample likelihoods, which relates to the approach of \glspl{VAE} \citep{kingma2013auto}. 
\citet{sanchez2019out} propose Eval\acrshort{GAN}, a method to estimate the likelihood. Given an observation $\bx^\star$, Eval\acrshort{GAN} first finds the closest observation $\widetilde{\bx}$ that the \gls{GAN} is able to generate, and then it estimates the likelihood $p(\bx^\star)$ by approximating the proportion of samples $\bz\sim p(\bz)$ that lead to samples $\bx$ that are close to $\widetilde{\bx}$. Eval\gls{GAN} requires selecting an appropriate distance metric for each problem and evaluates \glspl{GAN} trained with the usual implicit generative distribution. Finally, \citet{grover2018flow} assume invertibility of the generator to make log-likelihood tractable. 
 
\section{Empirical Study}\label{sec:empirical}

Here we demonstrate Pres\glspl{GAN}' ability to prevent mode collapse and generate high-quality samples. We also evaluate its predictive performance as measured by log-likelihood.

\subsection{An Illustrative Example}
\label{subsec:illustrative}

In this section, we fit a \gls{GAN} to a toy synthetic dataset of $10$ modes. We choose the hyperparameters such that the \gls{GAN} collapses. We then apply these same hyperparameters to fit a Pres\gls{GAN} on the same synthetic dataset. This experiment demonstrates the Pres\gls{GAN}'s ability to correct the mode collapse problem of a \gls{GAN}.

We form the target distribution by organizing a uniform mixture of $K=10$ two-dimensional Gaussians on a ring. The radius of the ring is $r = 3$ and each Gaussian has standard deviation $0.05$. We then slice the circle 
into $K$ parts. The location of the centers of the mixture components are determined as follows. Consider the $k^{\textrm{th}}$ mixture component. Its coordinates in the $2$D space are
\begin{align*}
	\text{center}_x &=  r \cdot \text{cos}\Big(k \cdot \frac{2\pi}{K}\Big) \quad \text{and} \quad
	\text{center}_y = r \cdot \text{sin}\Big(k \cdot \frac{2\pi}{K}\Big)
	.
\end{align*}
We draw $5{,}000$ samples from the target distribution and fit a \gls{GAN} and a Pres\gls{GAN}.

We set the dimension of the latent variables $\bz$ used as the input to the generators to $10$. We let both the generators and the discriminators have three fully connected layers with tanh activations and $128$ hidden units in each layer. We set the minibatch size to $100$ and use Adam for optimization \citep{kingma2014adam}, with a learning rate of $10^{-3}$ and $10^{-4}$ for the discriminator and the generator respectively. The Adam hyperparameters are $\beta_1 = 0.5$ and $\beta_2 = 0.999$. We take one step to optimize the generator for each step of the discriminator. We pick a random minibatch at each iteration and run both the \gls{GAN} and the Pres\gls{GAN} for $500$ epochs. 

For Pres\gls{GAN} we set the burn-in and the number of \gls{HMC} samples to $2$. We choose a standard number of $5$ leapfrog steps and set the \gls{HMC} learning rate to $0.02$. The acceptance rate is fixed at $0.67$. The log-variance of the noise of the generative distribution of Pres\gls{GAN} is initialized at $0.0$. We put a threshold on the variance to a minimum value of $\bsigma_{\text{low}} = 10^{-2}$ and a maximum value of $\bsigma_{\text{high}} = 0.3$. The regularization parameter $\lambda$ is $0.1$. We fit the log-variance using Adam with a learning rate of $10^{-4}$. 

Figure\nobreakspace \ref {fig:toy_example_fig1} demonstrates how the Pres\gls{GAN} alleviates mode collapse. The distribution learned by the regular \gls{GAN} misses $4$ modes of the target distribution. The Pres\gls{GAN} is able to recover all the modes of the target distribution. 

\subsection{Assessing mode collapse}
\label{subsec:assessing_mode_collapse}

In this section we evaluate Pres\glspl{GAN}' ability to mitigate mode collapse on real datasets. We run two sets of experiments. In the first set of experiments we adopt the current experimental protocol for assessing mode collapse in the \gls{GAN} literature. That is, we use the \textsc{mnist} and \textsc{stackedmnist} datasets, for which we know the true number of modes, and report two metrics: the number of modes recovered by the Pres\gls{GAN} and the \gls{KL} divergence between the label distribution induced by the Pres\gls{GAN} and the true label distribution. In the second set of experiments we demonstrate that mode collapse can happen in \glspl{GAN} even when the number of modes is as low as $10$ but the data is imbalanced. 

\parhead{Increased number of modes.} We consider the \textsc{mnist} and \textsc{stackedmnist} datasets. \textsc{mnist} is a dataset of hand-written digits,\footnote{See \url{http://yann.lecun.com/exdb/mnist}.} in which each $28\times 28 \times 1$ image corresponds to a digit. There are $60{,}000$ training digits and $10{,}000$ digits in the test set. \textsc{mnist} has $10$ modes, one for each digit. \textsc{stackedmnist} is formed by concatenating triplets of randomly chosen \textsc{mnist} digits along the color channel to form images of size $28\times 28\times 3$ \citep{Metz2017}. We keep the same size as the original \textsc{mnist}, $60{,}000$ training digits for $10{,}000$ test digits. The total number of modes in \textsc{stackedmnist} is $1{,}000$, corresponding to the number of possible triplets.

We consider \gls{DCGAN} as the base architecture and, following \citet{radford2015unsupervised}, we resize the spatial resolution of images to $64\times 64$ pixels.

\begin{table*}[t]
	\centering
	\small
	\captionof{table}{Assessing mode collapse on \textsc{mnist}. The true total number of modes is $10$. All methods capture all the $10$ modes. The \acrshort{KL} captures a notion of discrepancy between the labels of real versus generated images. Pres\acrshort{GAN} generates images whose distribution of labels is closer to the data distribution, as evidenced by lower \acrshort{KL} scores.}
	\begin{tabular}{ccc}
	\toprule
	 Method & Modes & KL \\
	 \hline
	 \acrshort{DCGAN} \citep{radford2015unsupervised} &  $10 \pm 0.0$ & $0.902 \pm 0.036$ \\
	 \acrshort{VEEGAN} \citep{srivastava2017veegan} & $10 \pm 0.0$ & $0.523\pm0.008$  \\
	  \acrshort{PACGAN} \citep{lin2018pacgan} & $10 \pm 0.0$ & $0.441\pm0.009$ \\
	  Pres\acrshort{GAN} (this paper) & $\textbf{10} \pm \textbf{0.0}$ & $\mathbf{0.003 \pm 0.001}$ \\
	\bottomrule
	\end{tabular}
	\label{tab:collapse_dimensionality_mnist}
\end{table*}

\begin{table*}[t]
	\centering
	\small
	\captionof{table}{Assessing mode collapse on \textsc{stackedmnist}. The true total number of modes is $1{,}000$. All methods suffer from collapse except Pres\gls{GAN}, which captures nearly all the modes of the data distribution. Furthermore, Pres\acrshort{GAN} generates images whose distribution of labels is closer to the data distribution, as evidenced by lower \acrshort{KL} scores.}
	\begin{tabular}{ccc}
	\toprule
	 Method & Modes & KL \\
	 \hline
	 \acrshort{DCGAN} \citep{radford2015unsupervised} &  $392.0 \pm 7.376$ & $8.012 \pm 0.056$ \\
	 \acrshort{VEEGAN} \citep{srivastava2017veegan} & $761.8\pm5.741$ & $2.173\pm0.045$ \\
	  \acrshort{PACGAN} \citep{lin2018pacgan} &  $992.0\pm1.673$ & $0.277\pm0.005$ \\
	  Pres\acrshort{GAN} (this paper) & $\mathbf{999.6\pm0.489}$ & $\mathbf{0.115}\pm\mathbf{0.007}$ \\
	\bottomrule
	\end{tabular}
	\label{tab:collapse_dimensionality_smnist}
\end{table*}

To measure the degree of mode collapse we form two diversity metrics, following \citet{srivastava2017veegan}. Both of these metrics require to fit a classifier to the training data. Once the classifier has been fit, we sample $S$ images from the generator. The first diversity metric is the \emph{number of modes captured}, measured by the number of classes that are captured by the classifier. We say that a class $k$ has been captured if there is at least one generated sample for which the probability of being assigned to class $k$ is the largest. The second diversity metric is the \emph{\gls{KL} divergence} between two discrete distributions: the empirical average of the (soft) output of the classifier on generated images, and the empirical average of the (soft) output of the classifier on real images from the test set. We choose the number of generated images $S$ to match the number of test samples on each dataset. That is, $S=10{,}000$ for both \textsc{mnist} and \textsc{stackedmnist}. We expect the \gls{KL} divergence to be zero if the distribution of the generated samples is indistinguishable from that of the test samples. 

We measure the two mode collapse metrics described above against \gls{DCGAN} \citep{radford2015unsupervised} (the base architecture of Pres\gls{GAN} for this experiment). 
We also compare against other methods that aim at alleviating mode collapse in \glspl{GAN}, namely, \acrshort{VEEGAN} \citep{srivastava2017veegan} and \acrshort{PACGAN} \citep{lin2018pacgan}. 
For Pres\gls{GAN} we set the entropy regularization parameter $\lambda$ to $0.01$. 
We chose the variance thresholds to be $\bsigma_{\text{low}} = 0.001$ and $\bsigma_{\text{high}} = 0.3$.

\begin{table*}[t]
	\centering
	\small
	\captionof{table}{Assessing the impact of the entropy regularization parameter $\lambda$ on mode collapse on \textsc{mnist} and \textsc{stackedmnist}. When $\lambda = 0$ (i.e., no entropy regularization is applied to the generator), then mode collapse occurs as expected. When entropy regularization is applied but the value of $\lambda$ is very small ($\lambda = 10^{-6}$) then mode collapse can still occur as the level of regularization is not enough. When the value of $\lambda$ is appropriate for the data then mode collapse does not occur. Finally, when $\lambda$ is too high then mode collapse can occur because the entropy maximization term dominates and the data is poorly fit.}
	\begin{tabular}{ccccc}
	\toprule
	   &  \multicolumn{2}{c}{\textsc{mnist}} &  \multicolumn{2}{c}{\textsc{stackedmnist}} \\
	 \midrule
	 $\lambda$ & Modes & KL & Modes & KL \\
	 \hline
	$0$ & $10 \pm 0.0$ & $0.050 \pm 0.0035$ & $418.2 \pm 7.68$ & $4.151 \pm 0.0296$ \\
	$10^{-6}$ & $10\pm0.0$ & $0.005\pm0.0008$ & $989.8\pm1.72$& $0.239\pm0.0059$ \\
	$10^{-2}$ & $\textbf{10} \pm \textbf{0.0}$ & $\textbf{0.003}\pm\textbf{0.0006}$ & $\textbf{999.6}\pm\textbf{0.49}$& $0.115\pm0.0074$ \\
	$5\times 10^{-2}$ & $10 \pm 0.0$ & $0.004\pm0.0008$ & $999.4\pm0.49$ & $\textbf{0.099}\pm\textbf{0.0047}$ \\
	$10^{-1}$ & $10 \pm 0.0$ & $0.005\pm0.0004$ & $999.4\pm0.80$ & $0.102\pm0.0032$\\
	$5\times 10^{-1}$ & $10\pm0.0$ & $0.006\pm0.0011$ & $907.0\pm9.27$& $0.831\pm0.0209$ \\
	\bottomrule
	\end{tabular}
	\label{tab:collapse_dim_bis}
\end{table*}

\begin{figure*}[t]
	\centering
	\vspace*{-10pt}
	\centerline{\includegraphics[width=1.2\textwidth]{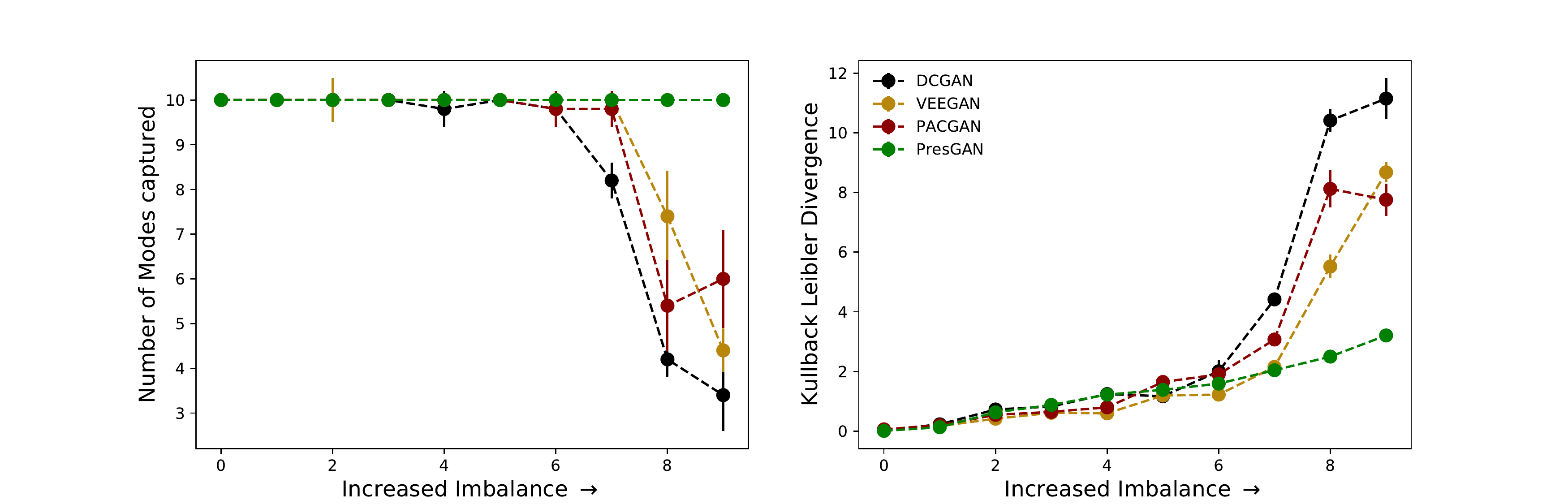}}
	\caption{Assessing mode collapse under increased data imbalance on \textsc{mnist}. The figures show the number of modes captured (higher is better) and the \gls{KL} divergence (lower is better) under increasingly imbalanced settings. The maximum number of modes in each case is $10$. All methods suffer from mode collapse as the level of imbalance increases except for the Pres\gls{GAN} which is robust to data imbalance.}
	\label{fig:imbalance}
	\vspace*{-8pt}
\end{figure*}

Tables\nobreakspace \ref {tab:collapse_dimensionality_mnist} and\nobreakspace  \ref {tab:collapse_dimensionality_smnist} show the number of captured modes and the \gls{KL} for each method. The results are averaged across $5$ runs. All methods capture all the modes of \textsc{mnist}. This is not the case on \textsc{stackedmnist}, where the Pres\gls{GAN} is the only method that can capture all the modes. 
Finally, the proportion of observations in each mode of Pres\gls{GAN} is closer to the true proportion in the data, as evidenced by lower \acrshort{KL} divergence scores.

We also study the impact of the entropy regularization by varying the hyperparameter $\lambda$ from $0$ to $0.5$. Table\nobreakspace \ref {tab:collapse_dim_bis} illustrates the results. Unsurprisingly, when there is no entropy regularization, i.e., when $\lambda = 0$, then mode collapse occurs. This is also the case when the level of regularization is not enough ($\lambda = 10^{-6}$). There is a whole range of values for $\lambda$ such that mode collapse does not occur ($\lambda \in \{0.01, 0.05, 0.1\}$). Finally, when $\lambda$ is too high for the data and architecture under study, mode collapse can still occur. This is because when $\lambda$ is too high, the entropy regularization term dominates the loss in Eq.\nobreakspace \ref {eq:sigan_loss} and in turn the generator does not fit the data as well. This is also evidenced by the higher \gls{KL} divergence score when $\lambda = 0.5$ vs.\ when $0 < \lambda < 0.5$.

\parhead{Increased data imbalance.} We now show that mode collapse can occur in \glspl{GAN} when the data is imbalanced, even when the number of modes of the data distribution is small. We follow \citet{dieng2018learning} and consider a perfectly balanced version of \textsc{mnist} as well as nine imbalanced versions. To construct the balanced dataset we used $5{,}000$ training examples per class, totaling $50{,}000$ training examples. We refer to this original balanced dataset as ${D_0}$. Each additional training set ${D_k}$ leaves only $5$ training examples for each class $j \leq k$, and $5{,}000$ for the rest. (See the Appendix for all the class distributions.)

We used the same classifier trained on the unmodified \textsc{mnist} but fit each method on each of the $9$ new \textsc{mnist} distributions. We chose $\lambda = 0.1$ for Pres\gls{GAN}. Figure\nobreakspace \ref {fig:imbalance} illustrates the results in terms of both metrics---number of modes and \gls{KL} divergence. \gls{DCGAN}, \acrshort{VEEGAN}, and \acrshort{PACGAN} face mode collapse as the level of imbalance increases. This is not the case for Pres\gls{GAN}, which is robust to imbalance and captures all the $10$ modes. 

\subsection{Assessing sample quality}
\label{subsec:assessing_sample_quality}

In this section we assess Pres\glspl{GAN}' ability to generate samples of high perceptual quality. We rely on perceptual quality of generated samples and on \gls{FID} scores \citep{heusel2017gans}. We also consider two different \gls{GAN} architectures, the standard \gls{DCGAN} and the more recent Style\gls{GAN}, to show robustness of Pres\glspl{GAN} vis-a-vis the underlying \gls{GAN} architecture.

\parhead{\gls{DCGAN}.} We use \gls{DCGAN} \citep{radford2015unsupervised} as the base architecture and build Pres\gls{GAN} on top of it. We consider four datasets: \textsc{mnist}, \textsc{stackedmnist}, \textsc{cifar}-10, and CelebA. \textsc{cifar}-10 \citep{krizhevsky2009learning} is a well-studied dataset of $32 \times 32$ images that are classified into one of the following categories: airplane, automobile, bird, cat, deer, dog, frog, horse, ship, and truck. CelebA \citep{liu2015deep} is a large-scale face attributes dataset. Following \citet{radford2015unsupervised}, we resize all images to $64 \times 64$ pixels.
We use the default \gls{DCGAN} settings. We refer the reader to the code we used for \gls{DCGAN}, which was taken from \url{https://github.com/pytorch/examples/tree/master/dcgan}. We set the seed to $2019$ for reproducibility.

\begin{table*}[t]
	\centering
	\small
	\captionof{table}{\acrfull{FID} (lower is better). Pres\acrshort{GAN} has lower \acrshort{FID} scores than \acrshort{DCGAN}, \acrshort{VEEGAN}, and \acrshort{PACGAN}. This is because Pres\acrshort{GAN} mitigates mode collapse while preserving sample quality.}
	\begin{tabular}{ccc}
	\toprule
	 Method & Dataset &  \acrshort{FID} \\
	 \hline
	 \acrshort{DCGAN} \citep{radford2015unsupervised} & \textsc{mnist} & $113.129\pm0.490$ \\
	 \acrshort{VEEGAN} \citep{srivastava2017veegan} & \textsc{mnist} & $68.749\pm0.428$ \\
	  \acrshort{PACGAN}   \citep{lin2018pacgan}  & \textsc{mnist} & $58.535\pm0.135$ \\
	  Pres\acrshort{GAN} (this paper) & \textsc{mnist} & $\textbf{42.019}\pm\textbf{0.244}$ \\
	  \hline
	  \acrshort{DCGAN}  & \textsc{stackedmnist} & $97.788\pm0.199$ \\
	 \acrshort{VEEGAN} & \textsc{stackedmnist} &  $86.689\pm0.194$ \\
	  \acrshort{PACGAN} & \textsc{stackedmnist} & $117.128\pm0.172$ \\
	  Pres\acrshort{GAN} & \textsc{stackedmnist} & $\textbf{23.965}\pm\textbf{0.134}$ \\
	  \hline
	  \acrshort{DCGAN}  & \textsc{cifar}-10 & $103.049\pm0.195$  \\
	 \acrshort{VEEGAN} & \textsc{cifar}-10 & $95.181\pm0.416$  \\
	  \acrshort{PACGAN} & \textsc{cifar}-10 &  $54.498\pm0.337$ \\
	  Pres\acrshort{GAN} & \textsc{cifar}-10 & $\textbf{52.202}\pm\textbf{0.124}$ \\
	  \hline
	  \acrshort{DCGAN}  & \textsc{celeba} & $39.001\pm0.243$ \\
	 \acrshort{VEEGAN} & \textsc{celeba} & $46.188\pm0.229$  \\
	  \acrshort{PACGAN} & \textsc{celeba} & $36.058\pm0.212$ \\
	  Pres\acrshort{GAN} & \textsc{celeba} & $\textbf{29.115}\pm\textbf{0.218}$ \\
	\bottomrule
	\end{tabular}
	\label{tab:fid}
\end{table*}

There are hyperparameters specific to Pres\gls{GAN}. These are the noise and \gls{HMC} hyperparameters. We set the learning rate for the noise parameters $\bsigma$ to $10^{-3}$ and constrain its values to be between $10^{-3}$ and $0.3$ for all datasets. We initialize $\log \bsigma$ to $-0.5$. We set the burn-in and the number of \gls{HMC} samples to $2$. We choose a standard number of $5$ leapfrog steps and set the \gls{HMC} learning rate to $0.02$. The acceptance rate is fixed at $0.67$. We found that different $\lambda$ values worked better for different datasets. We used $\lambda = 5\times 10^{-4}$ for \textsc{cifar}-10 and \textsc{celeba} $\lambda = 0.01$ for \textsc{mnist} and \textsc{stackedmnist}.

We found the Pres\gls{GAN}'s performance to be robust to the default settings for most of these hyperparameters. However we found the initialization for $\bsigma$ and its learning rate to play a role in the quality of the generated samples. The hyperparameters mentioned above for $\bsigma$ worked well for all datasets. 

Table\nobreakspace \ref {tab:fid} shows the \gls{FID} scores for \gls{DCGAN} and Pres\gls{GAN} across the four datasets.  
We can conclude that Pres\gls{GAN} generates images of high visual quality. In addition, the \gls{FID} scores are lower because Pres\gls{GAN} explores more modes than \gls{DCGAN}. Indeed, when the generated images account for more modes, the \gls{FID} sufficient statistics (the mean and covariance of the Inception-v3 pool3 layer) of the generated data get closer to the sufficient statistics of the empirical data distribution.

We also report the \gls{FID} for \acrshort{VEEGAN} and  \acrshort{PACGAN} in Table\nobreakspace \ref {tab:fid}.
\acrshort{VEEGAN} achieves better \gls{FID} scores than \acrshort{DCGAN} on all datasets but \textsc{celeba}. This is because \acrshort{VEEGAN} collapses less than \acrshort{DCGAN} as evidenced by Table\nobreakspace \ref {tab:collapse_dimensionality_mnist} and Table\nobreakspace \ref {tab:collapse_dimensionality_smnist}.  \acrshort{PACGAN} achieves better \gls{FID} scores than both \acrshort{DCGAN} and \acrshort{VEEGAN} on all datasets but on \textsc{stackedmnist} where it achieves a significantly worse \gls{FID} score. Finally, Pres\gls{GAN} outperforms all of these methods on the \gls{FID} metric on all datasets signaling its ability to mitigate mode collapse while preserving sample quality.

Besides the \gls{FID} scores, we also assess the visual quality of the generated images. In Section\nobreakspace \ref {app:samples} of the appendix, we show randomly generated (not cherry-picked) images from \gls{DCGAN},  \acrshort{VEEGAN}, \acrshort{PACGAN}, and Pres\gls{GAN}.  
For Pres\gls{GAN}, we show the mean of the conditional distribution of $\bx$ given $\bz$. The samples generated by Pres\gls{GAN} have high visual quality; in fact their quality is comparable to or better than the \gls{DCGAN} samples.

\parhead{Style\gls{GAN}.} We now consider a more recent \gls{GAN} architecture (Style\gls{GAN})~\citep{karras2019style} and a higher resolution image dataset (\textsc{ffhq}). 
\textsc{ffhq} is a diverse dataset of faces from Flickr\footnote{See \url{https://github.com/NVlabs/ffhq-dataset}.} introduced by \citet{karras2019style}. The dataset contains $70{,}000$ high-quality \textsc{png} images with considerable variation in terms of age, ethnicity, and image background. We use a resolution of $128 \times 128$ pixels. 

\begin{figure}[t]
	\includegraphics[width=0.45\textwidth]{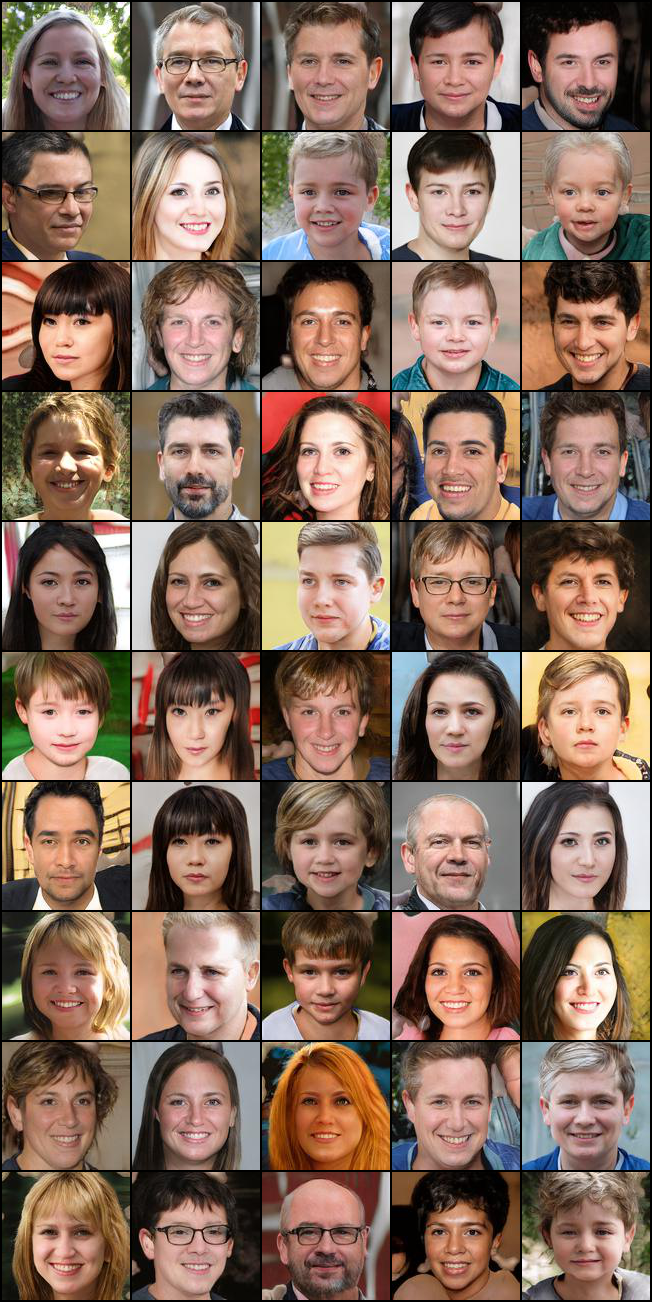}\quad\quad\quad
	\includegraphics[width=0.45\textwidth]{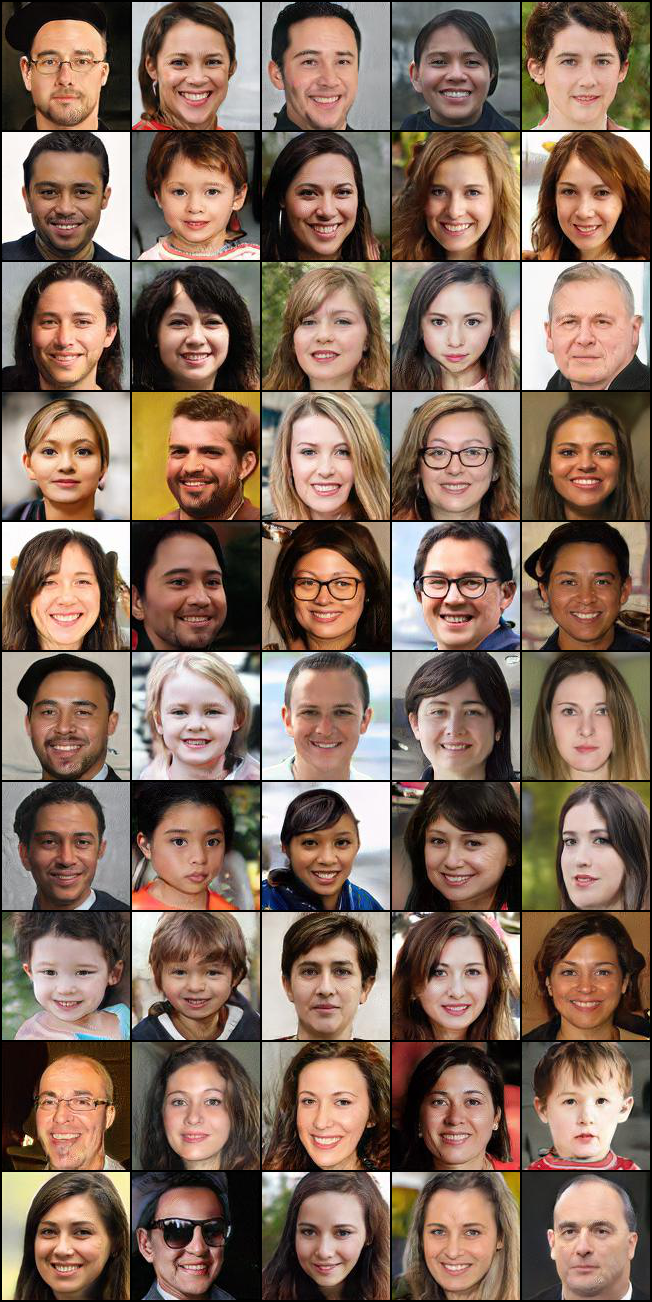}
	\caption{Generated images on \textsc{ffhq} for Style\gls{GAN} (left) and Pres\gls{GAN} (right). The Pres\gls{GAN} maintains the high perceptual quality of the Style\gls{GAN}.}
	\label{fig:ffhq}
\end{figure}

Style\gls{GAN} feeds multiple sources of noise $\bz$ to the generator. In particular, it adds Gaussian noise after each convolutional layer before evaluating the nonlinearity. Building Pres\gls{GAN} on top of Style\gls{GAN} therefore requires to sample all noise variables $\bz$ through \gls{HMC} at each training step. To speed up the training procedure, we only sample the noise variables corresponding to the input latent code and condition on all the other Gaussian noise variables. In addition, we do not follow the progressive growing of the networks of \citet{karras2019style} for simplicity.

For this experiment, we choose the same \gls{HMC} hyperparameters as for the previous experiments but restrict the variance of the generative distribution to be $\bsigma_{\text{high}} = 0.2$. We set $\lambda=0.001$ for this experiment. 

Figure\nobreakspace \ref {fig:ffhq} shows cherry-picked images generated from Style\gls{GAN} and Pres\gls{GAN}. We can observe that the Pres\gls{GAN} maintains as good perceptual quality as the base architecture. In addition, we also observed that the Style\gls{GAN} tends to produce some redundant images (these are not shown in Figure\nobreakspace \ref {fig:ffhq}), something that we did not observe with the Pres\gls{GAN}. This lack of diversity was also reflected in the \gls{FID} scores which were $14.72 \pm 0.09$ for Style\gls{GAN} and $12.15 \pm 0.09$ for Pres\gls{GAN}. These results suggest that entropy regularization effectively reduces mode collapse while preserving sample quality.

\subsection{Assessing held-out predictive log-likelihood}
\label{subsec:assessing_log_lik}

In this section we evaluate Pres\glspl{GAN} for generalization using predictive log-likelihood. We use the \gls{DCGAN} architecture to build Pres\gls{GAN} and evaluate the log-likelihood on two benchmark datasets, \textsc{mnist} and \textsc{cifar}-10. We use images of size $32 \times 32$.

We compare the generalization performance of the Pres\gls{GAN} against the \gls{VAE} \citep{kingma2013auto, rezende2014stochastic} by controlling for the architecture and the evaluation procedure. In particular, we fit a \gls{VAE} that has the same decoder architecture as the Pres\gls{GAN}. We form the \gls{VAE} encoder by using the same architecture as the \gls{DCGAN} discriminator and getting rid of the output layer. We used linear maps to get the mean and the log-variance of the approximate posterior. 

To measure how Pres\glspl{GAN} compare to traditional \glspl{GAN} in terms of log-likelihood, we also fit a Pres\gls{GAN} with $\lambda = 0$. 

\parhead{Evaluation.}
We control for the evaluation procedure and follow what's described in Section\nobreakspace \ref {subsec:loglik_evaluation} for all methods. We use $S=2{,}000$ samples to form the importance sampling estimator. Since the pixel values are normalized in $[-1, +1]$, we use a truncated Gaussian likelihood for evaluation. Specifically, for each pixel of the test image, we divide the Gaussian likelihood by the probability (under the generative model) that the pixel is within the interval $[-1, +1]$. We use the truncated Gaussian likelihood at test time only.

\parhead{Settings.}
For the Pres\gls{GAN}, we use the same \gls{HMC} hyperparameters as for the previous experiments. We constrain the variance of the generative distribution using $\bsigma_{\text{low}} = 0.001$ and $\bsigma_{\text{high}} = 0.2$. We use the default \gls{DCGAN} values for the remaining hyperparameters, including the optimization settings. For the \textsc{cifar}-10 experiment, we choose $\lambda = 0.001$. We set all learning rates to $0.0002$. We set the dimension of the latent variables to $100$. We ran both the \gls{VAE} and the Pres\gls{GAN} for a maximum of $200$ epochs. For \textsc{mnist}, we use the same settings as for \textsc{cifar}-10 but use $\lambda = 0.0001$ and ran all methods for a maximum of $50$ epochs.

\begin{table}[t]
	\centering
	\small
	\captionof{table}{Generalization performance as measured by negative log-likelihood (lower is better) on \textsc{mnist} and \textsc{cifar}-10. Here the \gls{GAN} denotes a Pres\gls{GAN} fitted without entropy regularization ($\lambda = 0$). The Pres\gls{GAN} reduces the gap in performance between the \gls{GAN} and the \gls{VAE} on both datasets.}
	\begin{tabular}{ccccc}
		\toprule
		& \multicolumn{2}{c}{\textsc{mnist}} & \multicolumn{2}{c}{\textsc{cifar}-10} \\
		& Train & Test & Train & Test \\ \midrule
		\gls{VAE}     & $-3483.94$ & $-3408.16$ & $-1978.91$ & $-1665.84$   \\
		\gls{GAN} & $-1410.78$ & $-1423.39$ & $-572.25$ & $-569.17$  \\
		Pres\gls{GAN} & $-1418.91$ & $-1432.50$ & $-1050.16$ & $-1031.70$  \\
		\bottomrule
	\end{tabular}
	\label{tab:loglik}
\end{table}

\parhead{Results.}
Table\nobreakspace \ref {tab:loglik} summarizes the results. Here \gls{GAN} denotes the Pres\gls{GAN} fitted using $\lambda = 0$. The \gls{VAE} outperforms both the \gls{GAN} and the Pres\gls{GAN} on both \textsc{mnist} and \textsc{cifar}-10. This is unsurprising given \glspl{VAE} are fitted to maximize log-likelihood. The \gls{GAN}'s performance on \textsc{cifar}-10 is particularly bad, suggesting it suffered from mode collapse. The Pres\gls{GAN}, which mitigates mode collapse achieves significantly better performance than the \gls{GAN} on \textsc{cifar}-10. To further analyze the generalization performance, we also report the log-likelihood on the training set in Table\nobreakspace \ref {tab:loglik}. We can observe that the difference between the training log-likelihood and the test log-likelihood is very small for all methods.
 
\section{Epilogue}
\label{sec:discussion}

We introduced the Pres\gls{GAN}, a variant of \glspl{GAN} that addresses two of their limitations. Pres\glspl{GAN} prevent mode collapse and are amenable to predictive log-likelihood evaluation. Pres\glspl{GAN} model data by adding noise to the output of a density network and optimize an entropy-regularized adversarial loss. The added noise stabilizes training, renders approximation of predictive log-likelihoods tractable, and enables unbiased estimators for the gradients of the entropy of the generative distribution. We evaluated Pres\glspl{GAN} on several image datasets. We found they effectively prevent mode collapse and generate samples of high perceptual quality. We further found that Pres\glspl{GAN} reduce the gap in performance between \glspl{GAN} and \glspl{VAE} in terms of predictive log-likelihood.

We found the level of entropy regularization $\lambda$ plays an important role in mode collapse. We leave as future work the task of finding the optimal $\lambda$. 
We now discuss some insights that we concluded from our empirical study in Section\nobreakspace \ref {sec:empirical}.

\parhead{Implicit distributions and sample quality.} It's been traditionally observed that \glspl{GAN} generate samples with higher perceptual quality than \glspl{VAE}. This can be explained by looking at the two ways in which \glspl{GAN} and \glspl{VAE} differ; the generative distribution and the objective function. \glspl{VAE} use prescribed generative distributions and optimize likelihood whereas \glspl{GAN} use implicit generative distributions and optimize an adversarial loss. Our results in Section\nobreakspace \ref {sec:empirical} suggest that the implicit generators of traditional \glspl{GAN} are not the key to high sample quality; rather, the key is the adversarial loss. This is because Pres\glspl{GAN} use the same prescribed generative distributions as \glspl{VAE} and achieve similar or sometimes better sample quality than \glspl{GAN}. 

\parhead{Mode collapse, diversity, and imbalanced data.}
The current literature on measuring mode collapse in \glspl{GAN} only focuses on showing that mode collapse happens when the number of modes in the data distribution is high. 
Our results show that mode collapse can happen not only when the number of modes of the data distribution is high, but also when the data is imbalanced; even when the number of modes is low. 
Imbalanced data are ubiquitous. Therefore, mitigating mode collapse in \glspl{GAN} is important for the purpose of diverse data generation.

\parhead{\glspl{GAN} and generalization.}
The main method to evaluate generalization for density estimators is predictive log-likelihood. Our results agree with the current literature that \glspl{GAN} don't generalize as well as \glspl{VAE} which are specifically trained to maximize log-likelihood. However, our results show that entropy-regularized adversarial learning can reduce the gap in generalization performance between \glspl{GAN} and \glspl{VAE}. Methods that regularize \glspl{GAN} with the maximum likelihood objective achieve good generalization performance when compared to \glspl{VAE} but they sacrifice sample quality when doing so~\citep{grover2018flow}. In fact we also experienced this tension between sample quality and high log-likelihood in practice. 

Why is there such a gap in generalization, as measured by predictive log-likelihood, between \glspl{GAN} and \glspl{VAE}? In our empirical study in Section\nobreakspace \ref {sec:empirical} we controlled for the architecture and the evaluation procedure which left us to compare maximizing likelihood against adversarial learning. Our results suggest mode collapse alone does not explain the gap in generalization performance between \glspl{GAN} and \glspl{VAE}. Indeed Table\nobreakspace \ref {tab:loglik} shows that even on \textsc{MNIST}, where mode collapse does not happen, the \gls{VAE} achieves significantly better log-likelihood than a \gls{GAN}. 

We looked more closely at the encoder fitted at test time to evaluate log-likelihood for both the \gls{VAE} and the \gls{GAN} (not shown in this paper). We found that the encoder implied by a fitted \gls{GAN} is very underdispersed compared to the encoder implied by a fitted \gls{VAE}. Underdispersed proposals have a negative impact on importance sampling estimates of log-likelihood. We tried to produce a more overdispersed proposal using the procedure described in Section\nobreakspace \ref {subsec:loglik_evaluation}. However we leave as future work learning overdispersed proposals for \glspl{GAN} for the purpose of log-likelihood evaluation.

\subsection*{Acknowledgements}
We thank Ian Goodfellow, Andriy Mnih, Aaron Van den Oord, and Laurent Dinh for their comments. Francisco J.\ R.\ Ruiz is supported by the European Union's Horizon 2020 research and innovation programme under the Marie Sk\l{}odowska-Curie grant agreement No.\ 706760. Adji B.\ Dieng is supported by a Google PhD Fellowship.

\bibliographystyle{apa}
\bibliography{main}

\begin{thebibliography}{}

\bibitem[\protect\astroncite{Arjovsky et~al.}{2017}]{arjovsky2017wasserstein}
Arjovsky, M., Chintala, S., and Bottou, L. (2017).
\newblock Wasserstein generative adversarial networks.
\newblock In {\em International conference on machine learning}, pages
  214--223.

\bibitem[\protect\astroncite{Arora et~al.}{2017}]{arora2017generalization}
Arora, S., Ge, R., Liang, Y., Ma, T., and Zhang, Y. (2017).
\newblock Generalization and equilibrium in generative adversarial nets
  ({GANs}).
\newblock In {\em International Conference on Machine Learning}.

\bibitem[\protect\astroncite{Arora et~al.}{2018}]{arora2018gans}
Arora, S., Risteski, A., and Zhang, Y. (2018).
\newblock Do gans learn the distribution? some theory and empirics.

\bibitem[\protect\astroncite{Azadi et~al.}{2018}]{azadi2018discriminator}
Azadi, S., Olsson, C., Darrell, T., Goodfellow, I., and Odena, A. (2018).
\newblock Discriminator rejection sampling.
\newblock {\em arXiv preprint arXiv:1810.06758}.

\bibitem[\protect\astroncite{Balaji et~al.}{2018}]{balaji2018entropic}
Balaji, Y., Hassani, H., Chellappa, R., and Feizi, S. (2018).
\newblock Entropic {GANs} meet {VAEs}: A statistical approach to compute sample
  likelihoods in gans.
\newblock {\em arXiv preprint arXiv:1810.04147}.

\bibitem[\protect\astroncite{Belghazi et~al.}{2018}]{belghazi2018mine}
Belghazi, M.~I., Baratin, A., Rajeswar, S., Ozair, S., Bengio, Y., Courville,
  A., and Hjelm, R.~D. (2018).
\newblock Mine: mutual information neural estimation.
\newblock {\em arXiv preprint arXiv:1801.04062}.

\bibitem[\protect\astroncite{Bi{\'n}kowski
  et~al.}{2018}]{binkowski2018demystifying}
Bi{\'n}kowski, M., Sutherland, D.~J., Arbel, M., and Gretton, A. (2018).
\newblock Demystifying {MMD} {GANs}.
\newblock {\em arXiv:1801.01401}.

\bibitem[\protect\astroncite{Bishop}{2006}]{bishop2006pattern}
Bishop, C.~M. (2006).
\newblock {\em Pattern recognition and machine learning}.
\newblock springer.

\bibitem[\protect\astroncite{Blei et~al.}{2017}]{blei2017variational}
Blei, D.~M., Kucukelbir, A., and McAuliffe, J.~D. (2017).
\newblock Variational inference: A review for statisticians.
\newblock {\em Journal of the American Statistical Association},
  112(518):859--877.

\bibitem[\protect\astroncite{Bowman et~al.}{2015}]{bowman2015generating}
Bowman, S.~R., Vilnis, L., Vinyals, O., Dai, A.~M., Jozefowicz, R., and Bengio,
  S. (2015).
\newblock Generating sentences from a continuous space.
\newblock {\em arXiv:1511.06349}.

\bibitem[\protect\astroncite{Brock et~al.}{2018}]{brock2018large}
Brock, A., Donahue, J., and Simonyan, K. (2018).
\newblock Large scale gan training for high fidelity natural image synthesis.
\newblock {\em arXiv:1809.11096}.

\bibitem[\protect\astroncite{Cao et~al.}{2018}]{cao2018improving}
Cao, Y., Ding, G.~W., Lui, K. Y.-C., and Huang, R. (2018).
\newblock Improving gan training via binarized representation entropy (bre)
  regularization.
\newblock {\em arXiv preprint arXiv:1805.03644}.

\bibitem[\protect\astroncite{Dieng et~al.}{2018a}]{dieng2018learning}
Dieng, A.~B., Cho, K., Blei, D.~M., and LeCun, Y. (2018a).
\newblock Learning with reflective likelihoods.

\bibitem[\protect\astroncite{Dieng et~al.}{2018b}]{dieng2018avoiding}
Dieng, A.~B., Kim, Y., Rush, A.~M., and Blei, D.~M. (2018b).
\newblock Avoiding latent variable collapse with generative skip models.
\newblock {\em arXiv:1807.04863}.

\bibitem[\protect\astroncite{Dieng and Paisley}{2019}]{dieng2019reweighted}
Dieng, A.~B. and Paisley, J. (2019).
\newblock Reweighted expectation maximization.
\newblock {\em arXiv preprint arXiv:1906.05850}.

\bibitem[\protect\astroncite{Diggle and Gratton}{1984}]{diggle1984monte}
Diggle, P.~J. and Gratton, R.~J. (1984).
\newblock {M}onte {C}arlo methods of inference for implicit statistical models.
\newblock {\em Journal of the Royal Statistical Society: Series B
  (Methodological)}, 46(2):193--212.

\bibitem[\protect\astroncite{Donahue et~al.}{2016}]{donahue2016adversarial}
Donahue, J., Kr{\"a}henb{\"u}hl, P., and Darrell, T. (2016).
\newblock Adversarial feature learning.
\newblock {\em arXiv preprint arXiv:1605.09782}.

\bibitem[\protect\astroncite{Dumoulin et~al.}{2016}]{dumoulin2016adversarially}
Dumoulin, V., Belghazi, I., Poole, B., Mastropietro, O., Lamb, A., Arjovsky,
  M., and Courville, A. (2016).
\newblock Adversarially learned inference.
\newblock {\em arXiv preprint arXiv:1606.00704}.

\bibitem[\protect\astroncite{Dziugaite et~al.}{2015}]{dziugaite2015training}
Dziugaite, G.~K., Roy, D.~M., and Ghahramani, Z. (2015).
\newblock Training generative neural networks via maximum mean discrepancy
  optimization.
\newblock {\em arXiv preprint arXiv:1505.03906}.

\bibitem[\protect\astroncite{Freund and Schapire}{1997}]{freund1997decision}
Freund, Y. and Schapire, R.~E. (1997).
\newblock A decision-theoretic generalization of on-line learning and an
  application to boosting.
\newblock {\em Journal of computer and system sciences}, 55(1):119--139.

\bibitem[\protect\astroncite{Genevay et~al.}{2017}]{genevay2017learning}
Genevay, A., Peyr{\'e}, G., and Cuturi, M. (2017).
\newblock Learning generative models with sinkhorn divergences.
\newblock {\em arXiv preprint arXiv:1706.00292}.

\bibitem[\protect\astroncite{Goodfellow
  et~al.}{2014}]{goodfellow2014generative}
Goodfellow, I., Pouget-Abadie, J., Mirza, M., Xu, B., Warde-Farley, D., Ozair,
  S., Courville, A., and Bengio, Y. (2014).
\newblock Generative adversarial nets.
\newblock In {\em Advances in neural information processing systems}, pages
  2672--2680.

\bibitem[\protect\astroncite{Grover et~al.}{2018}]{grover2018flow}
Grover, A., Dhar, M., and Ermon, S. (2018).
\newblock Flow-gan: Combining maximum likelihood and adversarial learning in
  generative models.
\newblock In {\em Thirty-Second AAAI Conference on Artificial Intelligence}.

\bibitem[\protect\astroncite{Heusel et~al.}{2017}]{heusel2017gans}
Heusel, M., Ramsauer, H., Unterthiner, T., Nessler, B., and Hochreiter, S.
  (2017).
\newblock {GANs} trained by a two time-scale update rule converge to a local
  {N}ash equilibrium.
\newblock In {\em Advances in Neural Information Processing Systems}.

\bibitem[\protect\astroncite{Husz\'{a}r}{2016}]{ferenc2016instance}
Husz\'{a}r, F. (2016).
\newblock Instance noise: a trick for stabilising gan training.
\newblock
  \url{https://www.inference.vc/instance-noise-a-trick-for-stabilising-gan-training/}.

\bibitem[\protect\astroncite{Isola et~al.}{2017}]{isola2017image}
Isola, P., Zhu, J.-Y., Zhou, T., and Efros, A.~A. (2017).
\newblock Image-to-image translation with conditional adversarial networks.
\newblock In {\em Proceedings of the IEEE conference on computer vision and
  pattern recognition}, pages 1125--1134.

\bibitem[\protect\astroncite{Jaynes}{2003}]{jaynes2003probability}
Jaynes, E.~T. (2003).
\newblock {\em Probability theory: The logic of science}.
\newblock Cambridge university press.

\bibitem[\protect\astroncite{Jordan}{1998}]{jordan1998learning}
Jordan, M.~I. (1998).
\newblock {\em Learning in graphical models}, volume~89.
\newblock Springer Science \& Business Media.

\bibitem[\protect\astroncite{Karras et~al.}{2019}]{karras2019style}
Karras, T., Laine, S., and Aila, T. (2019).
\newblock A style-based generator architecture for generative adversarial
  networks.
\newblock In {\em Conference on Computer Vision and Pattern Recognition}.

\bibitem[\protect\astroncite{Kingma and Ba}{2014}]{kingma2014adam}
Kingma, D.~P. and Ba, J. (2014).
\newblock Adam: A method for stochastic optimization.
\newblock {\em arXiv preprint arXiv:1412.6980}.

\bibitem[\protect\astroncite{Kingma and Welling}{2013}]{kingma2013auto}
Kingma, D.~P. and Welling, M. (2013).
\newblock Auto-encoding variational bayes.
\newblock {\em arXiv preprint arXiv:1312.6114}.

\bibitem[\protect\astroncite{Krizhevsky et~al.}{2009}]{krizhevsky2009learning}
Krizhevsky, A., Hinton, G., et~al. (2009).
\newblock Learning multiple layers of features from tiny images.
\newblock Technical report, Citeseer.

\bibitem[\protect\astroncite{Kumar et~al.}{2019}]{kumar2019maximum}
Kumar, R., Goyal, A., Courville, A., and Bengio, Y. (2019).
\newblock Maximum entropy generators for energy-based models.
\newblock {\em arXiv preprint arXiv:1901.08508}.

\bibitem[\protect\astroncite{Ledig et~al.}{2017}]{ledig2017photo}
Ledig, C., Theis, L., Husz{\'a}r, F., Caballero, J., Cunningham, A., Acosta,
  A., Aitken, A., Tejani, A., Totz, J., Wang, Z., et~al. (2017).
\newblock Photo-realistic single image super-resolution using a generative
  adversarial network.
\newblock In {\em Proceedings of the IEEE conference on computer vision and
  pattern recognition}, pages 4681--4690.

\bibitem[\protect\astroncite{Li et~al.}{2015}]{li2015generative}
Li, Y., Swersky, K., and Zemel, R. (2015).
\newblock Generative moment matching networks.
\newblock In {\em International Conference on Machine Learning}, pages
  1718--1727.

\bibitem[\protect\astroncite{Lin et~al.}{2018}]{lin2018pacgan}
Lin, Z., Khetan, A., Fanti, G., and Oh, S. (2018).
\newblock Pacgan: The power of two samples in generative adversarial networks.
\newblock In {\em Advances in Neural Information Processing Systems}, pages
  1498--1507.

\bibitem[\protect\astroncite{Liu et~al.}{2015}]{liu2015deep}
Liu, Z., Luo, P., Wang, X., and Tang, X. (2015).
\newblock Deep learning face attributes in the wild.
\newblock In {\em Proceedings of the IEEE international conference on computer
  vision}, pages 3730--3738.

\bibitem[\protect\astroncite{MacKay}{1995}]{mackay1995bayesian}
MacKay, D.~J. (1995).
\newblock Bayesian neural networks and density networks.
\newblock {\em Nuclear Instruments and Methods in Physics Research Section A:
  Accelerators, Spectrometers, Detectors and Associated Equipment},
  354(1):73--80.

\bibitem[\protect\astroncite{Makhzani et~al.}{2015}]{makhzani2015adversarial}
Makhzani, A., Shlens, J., Jaitly, N., Goodfellow, I., and Frey, B. (2015).
\newblock Adversarial autoencoders.
\newblock {\em arXiv preprint arXiv:1511.05644}.

\bibitem[\protect\astroncite{Mescheder et~al.}{2017}]{mescheder2017adversarial}
Mescheder, L., Nowozin, S., and Geiger, A. (2017).
\newblock Adversarial variational bayes: Unifying variational autoencoders and
  generative adversarial networks.
\newblock In {\em Proceedings of the 34th International Conference on Machine
  Learning-Volume 70}, pages 2391--2400. JMLR. org.

\bibitem[\protect\astroncite{Metz et~al.}{2017}]{Metz2017}
Metz, L., Poole, B., Pfau, D., and Sohl-Dickstein, J. (2017).
\newblock Unrolled generative adversarial networks.
\newblock In {\em International Conference on Learning Representations}.

\bibitem[\protect\astroncite{Minka et~al.}{2005}]{minka2005divergence}
Minka, T. et~al. (2005).
\newblock Divergence measures and message passing.
\newblock Technical report, Technical report, Microsoft Research.

\bibitem[\protect\astroncite{Mnih et~al.}{2016}]{mnih2016asynchronous}
Mnih, V., Badia, A.~P., Mirza, M., Graves, A., Lillicrap, T., Harley, T.,
  Silver, D., and Kavukcuoglu, K. (2016).
\newblock Asynchronous methods for deep reinforcement learning.
\newblock In {\em International conference on machine learning}, pages
  1928--1937.

\bibitem[\protect\astroncite{Mohamed and
  Lakshminarayanan}{2016}]{mohamed2016learning}
Mohamed, S. and Lakshminarayanan, B. (2016).
\newblock Learning in implicit generative models.
\newblock {\em arXiv:1610.03483}.

\bibitem[\protect\astroncite{Neal}{2001}]{neal2001annealed}
Neal, R.~M. (2001).
\newblock Annealed importance sampling.
\newblock {\em Statistics and computing}, 11(2):125--139.

\bibitem[\protect\astroncite{Neal et~al.}{2011}]{neal2011mcmc}
Neal, R.~M. et~al. (2011).
\newblock Mcmc using hamiltonian dynamics.
\newblock {\em Handbook of markov chain monte carlo}, 2(11):2.

\bibitem[\protect\astroncite{Nowozin et~al.}{2016}]{nowozin2016f}
Nowozin, S., Cseke, B., and Tomioka, R. (2016).
\newblock f-gan: Training generative neural samplers using variational
  divergence minimization.
\newblock In {\em Advances in neural information processing systems}, pages
  271--279.

\bibitem[\protect\astroncite{Parzen}{1962}]{parzen1962estimation}
Parzen, E. (1962).
\newblock On estimation of a probability density function and mode.
\newblock {\em The annals of mathematical statistics}, 33(3):1065--1076.

\bibitem[\protect\astroncite{Radford et~al.}{2015}]{radford2015unsupervised}
Radford, A., Metz, L., and Chintala, S. (2015).
\newblock Unsupervised representation learning with deep convolutional
  generative adversarial networks.
\newblock {\em arXiv preprint arXiv:1511.06434}.

\bibitem[\protect\astroncite{Ravuri et~al.}{2018}]{ravuri2018learning}
Ravuri, S., Mohamed, S., Rosca, M., and Vinyals, O. (2018).
\newblock Learning implicit generative models with the method of learned
  moments.
\newblock {\em arXiv preprint arXiv:1806.11006}.

\bibitem[\protect\astroncite{Rezende et~al.}{2014}]{rezende2014stochastic}
Rezende, D.~J., Mohamed, S., and Wierstra, D. (2014).
\newblock Stochastic backpropagation and approximate inference in deep
  generative models.
\newblock {\em International Conference on Machine Learning}.

\bibitem[\protect\astroncite{Rigollet and Weed}{2018}]{rigollet2018entropic}
Rigollet, P. and Weed, J. (2018).
\newblock Entropic optimal transport is maximum-likelihood deconvolution.
\newblock {\em Comptes Rendus Math{\'e}matique}, 356(11-12):1228--1235.

\bibitem[\protect\astroncite{Rosca et~al.}{2017}]{rosca2017variational}
Rosca, M., Lakshminarayanan, B., Warde-Farley, D., and Mohamed, S. (2017).
\newblock Variational approaches for auto-encoding generative adversarial
  networks.
\newblock {\em arXiv:1706.04987}.

\bibitem[\protect\astroncite{Salimans et~al.}{2016}]{salimans2016improved}
Salimans, T., Goodfellow, I., Zaremba, W., Cheung, V., Radford, A., and Chen,
  X. (2016).
\newblock Improved techniques for training {GANs}.
\newblock In {\em Advances in neural information processing systems}.

\bibitem[\protect\astroncite{S{\'a}nchez-Mart{\'\i}n
  et~al.}{2019}]{sanchez2019out}
S{\'a}nchez-Mart{\'\i}n, P., Olmos, P.~M., and P{\'e}rez-Cruz, F. (2019).
\newblock Out-of-sample testing for gans.
\newblock {\em arXiv preprint arXiv:1901.09557}.

\bibitem[\protect\astroncite{Schulman et~al.}{2015}]{schulman2015trust}
Schulman, J., Levine, S., Abbeel, P., Jordan, M., and Moritz, P. (2015).
\newblock Trust region policy optimization.
\newblock In {\em International conference on machine learning}, pages
  1889--1897.

\bibitem[\protect\astroncite{S{\o}nderby et~al.}{2016}]{sonderby2016amortised}
S{\o}nderby, C.~K., Caballero, J., Theis, L., Shi, W., and Husz{\'a}r, F.
  (2016).
\newblock Amortised map inference for image super-resolution.
\newblock {\em arXiv preprint arXiv:1610.04490}.

\bibitem[\protect\astroncite{Soofi}{2000}]{soofi2000principal}
Soofi, E.~S. (2000).
\newblock Principal information theoretic approaches.
\newblock {\em Journal of the American Statistical Association},
  95(452):1349--1353.

\bibitem[\protect\astroncite{Srivastava et~al.}{2017}]{srivastava2017veegan}
Srivastava, A., Valkov, L., Russell, C., Gutmann, M.~U., and Sutton, C. (2017).
\newblock Veegan: Reducing mode collapse in gans using implicit variational
  learning.
\newblock In {\em Advances in Neural Information Processing Systems}, pages
  3308--3318.

\bibitem[\protect\astroncite{Titsias and
  L{\'a}zaro-Gredilla}{2014}]{titsias2014doubly}
Titsias, M. and L{\'a}zaro-Gredilla, M. (2014).
\newblock Doubly stochastic variational bayes for non-conjugate inference.
\newblock In {\em International conference on machine learning}, pages
  1971--1979.

\bibitem[\protect\astroncite{Titsias and Ruiz}{2018}]{titsias2018unbiased}
Titsias, M.~K. and Ruiz, F.~J. (2018).
\newblock Unbiased implicit variational inference.
\newblock {\em arXiv preprint arXiv:1808.02078}.

\bibitem[\protect\astroncite{Tolstikhin
  et~al.}{2017}]{tolstikhin2017wasserstein}
Tolstikhin, I., Bousquet, O., Gelly, S., and Schoelkopf, B. (2017).
\newblock Wasserstein auto-encoders.
\newblock {\em arXiv preprint arXiv:1711.01558}.

\bibitem[\protect\astroncite{Turner et~al.}{2018}]{turner2018metropolis}
Turner, R., Hung, J., Saatci, Y., and Yosinski, J. (2018).
\newblock Metropolis-hastings generative adversarial networks.
\newblock {\em arXiv preprint arXiv:1811.11357}.

\bibitem[\protect\astroncite{Ulyanov et~al.}{2018}]{ulyanov2018takes}
Ulyanov, D., Vedaldi, A., and Lempitsky, V. (2018).
\newblock It takes (only) two: Adversarial generator-encoder networks.
\newblock In {\em AAAI Conference on Artificial Intelligence}.

\bibitem[\protect\astroncite{Wainwright et~al.}{2008}]{wainwright2008graphical}
Wainwright, M.~J., Jordan, M.~I., et~al. (2008).
\newblock Graphical models, exponential families, and variational inference.
\newblock {\em Foundations and Trends{\textregistered} in Machine Learning},
  1(1--2):1--305.

\bibitem[\protect\astroncite{Wu et~al.}{2016}]{wu2016quantitative}
Wu, Y., Burda, Y., Salakhutdinov, R., and Grosse, R. (2016).
\newblock On the quantitative analysis of decoder-based generative models.
\newblock {\em arXiv preprint arXiv:1611.04273}.

\bibitem[\protect\astroncite{Xiao et~al.}{2018}]{xiao2018bourgan}
Xiao, C., Zhong, P., and Zheng, C. (2018).
\newblock {BourGAN}: Generative networks with metric embeddings.
\newblock In {\em Advances in Neural Information Processing Systems}.

\bibitem[\protect\astroncite{Yin and Zhou}{2019}]{yin2019semi}
Yin, M. and Zhou, M. (2019).
\newblock Semi-implicit generative model.
\newblock {\em arXiv preprint arXiv:1905.12659}.

\bibitem[\protect\astroncite{Zhang et~al.}{2017}]{zhang2017discrimination}
Zhang, P., Liu, Q., Zhou, D., Xu, T., and He, X. (2017).
\newblock On the discrimination-generalization tradeoff in gans.
\newblock {\em arXiv preprint arXiv:1711.02771}.

\end{thebibliography}

\section*{Appendix}
\stepcounter{section}

\subsection{Other Ways to Compute Predictive Log-Likelihood}\label{app:proposals}

Here we discuss different ways to obtain a proposal in order to approximate the predictive log-likelihood.
For a test instance $\bx^*$, we estimate the marginal log-likelihood $\log p_{\theta}(\bx^*)$ using importance sampling,
\begin{equation}\label{supp_eq:loglik}
	\log p_{\theta}(\bx^*) \approx \log \left( \frac{1}{S} \sum_{s=1}^{S} \frac{p_{\theta}\left(\bx^*\g \bz^{(s)}\right)\; p\left(\bz^{(s)}\right)}{r\left(\bz^{(s)}\g \bx^*\right)}\right),
\end{equation}
where we draw the $S$ samples $\bz^{(1)}, \dots, \bz^{(S)}$ from a proposal distribution $r(\bz\g \bx^*)$. We next discuss different ways to form the proposal $r(\bz\g \bx^*)$. 

One way to obtain the proposal is to set $r(\bz\g \bx^*)$ as a Gaussian distribution whose mean and variance are computed using samples from an \acrshort{HMC} algorithm with stationary distribution $p_{\theta}(\bz\g \bx^*)\propto p_{\theta}(\bx^*\g \bz)p(\bz)$. That is, the mean and variance of $r(\bz\g \bx^*)$ are set to the empirical mean and variance of the \acrshort{HMC} samples.

The procedure above requires to run an \acrshort{HMC} sampler, and thus it may be slow. We can accelerate the procedure with a better initialization of the \acrshort{HMC} chain.
Indeed, the second way to evaluate the log-likelihood also requires the \acrshort{HMC} sampler, but
it is initialized using a mapping $\bz = g_{\eta}(\bx^\star)$. 
The mapping $g_{\eta}(\bx^\star)$ is a network that maps from observed space $\bx$ to latent space $\bz$. The parameters $\eta$ of the network can be learned at test time using generated data. In particular, $\eta$ can be obtained by generating data from the fitted generator of Pres\acrshort{GAN} and then fitting $g_{\eta}(\bx^\star)$ to map $\bx$ to $\bz$ by maximum likelihood. This is, we first sample $M$ pairs $(\bz_m, \bx_m)_{m=1}^{M}$ from the learned generative distribution and then we obtain $\eta$ by minimizing
$\sum_{m=1}^{M} || \bz_m - g_{\eta}(\bx_m) ||_2^2$.
Once the mapping is fitted, we use it to initialize the \acrshort{HMC} chain.

A third way to obtain the proposal is to learn an encoder network $q_{\eta}(\bz \g \bx)$ jointly with the rest of the Pres\acrshort{GAN} parameters. This is effectively done by letting the discriminator distinguish between pairs $(\bx, \bz) \sim p_d(\bx)\cdot q_{\eta}(\bz \g \bx)$ and $(\bx, \bz) \sim p_{\theta}(\bx, \bz)$ rather than discriminate $\bx$ against samples from the generative distribution. These types of discriminator networks have been used to learn a richer latent space for \acrshort{GAN}~\citep{donahue2016adversarial, dumoulin2016adversarially}. 
In such cases, we can use the encoder network $q_{\eta}(\bz\g \bx)$ to define the proposal, either by setting $r(\bz\g \bx^*)=q_{\eta}(\bz\g \bx^*)$ or by initializing the \acrshort{HMC} sampler at the encoder mean.

The use of an encoder network is appealing but it requires a discriminator that takes pairs $(\bx,\bz)$. The approach that we follow in the paper also uses an encoder network but keeps the discriminator the same as for the base \acrshort{DCGAN}. We found this approach to work better in practice. More in detail, we use an encoder network $q_{\eta}(\bz \g \bx)$; however the encoder is fitted at test time by maximizing the variational \acrshort{ELBO}, given by $\sum_n \E{q_{\eta}(\bz_n\g \bx_n)}{\log p_{\theta}(\bx_n,\bz_n) - \log q_{\eta}(\bz_n\g \bx_n)}$. We set the proposal $r(\bz\g \bx^*)=q_{\eta}(\bz\g \bx^*)$. (Alternatively, the encoder can be used to initialize a sampler.)

\subsection{Assessing mode collapse under increased data imbalance}
In the main paper we show that mode collapse can happen not only when there are increasing number of modes, as done in the \gls{GAN} literature, but also when the data is imbalanced. We consider a perfectly balanced version of \textsc{mnist} by using 5,000 training examples per class, totalling 50,000 training examples. We refer to this original balanced dataset as {\bf D$1$}. We build nine additional training sets from this balanced dataset. Each additional training set {\bf D$k$} leaves only $5$ training examples for each class $j < k$. See Table~\ref{supp_tab:class_dist} for all the class distributions. 

\begin{table*}[!hbpt]
\centering
\caption{Class distributions using the \textsc{mnist} dataset. There are $10$ class---one class for each of 
the $10$ digits in \textsc{mnist}. The distribution D$1$ is uniform and the other distributions correspond 
to different imbalance settings as given by the proportions in the table. Note these proportions might not sum to one exactly because of rounding.}
\begin{tabular}[\textwidth]{ccccccccccc}
\toprule
Dist & $0$ & $1$ & $2$  & $3$ & $4$ &  $5$  &  $6$  &  $7$  & $8$  &  $9$ \\
\midrule
D$1$ &  $0.1$ & $0.1$ & $0.1$ & $0.1$ & $0.1$ & $0.1$ & $0.1$ & $0.1$ & $0.1$ & $0.1$ \\
D$2$ &  $10^{-3}$ & $0.11$ & $0.11$ & $0.11$ & $0.11$ & $0.11$ & $0.11$ & $0.11$ & $0.11$ & $0.11$   \\
D$3$ & $10^{-3}$ & $10^{-3}$ & $0.12$  & $0.12$  & $0.12$  & $0.12$  & $0.12$  & $0.12$  & $0.12$ & $0.12$     \\
D$4$ &  $10^{-3}$ & $10^{-3}$ & $10^{-3}$ & $0.14$  & $0.14$   & $0.14$  & $0.14$   & $0.14$   & $0.14$  & $0.14$  \\
D$5$ &  $10^{-3}$ & $10^{-3}$ & $10^{-3}$ & $10^{-3}$  & $0.17$ & $0.17$ & $0.17$ & $0.17$ & $0.17$ & $0.17$   \\
D$6$ &  $10^{-3}$ &$10^{-3}$ &  $10^{-3}$ & $10^{-3}$ & $10^{-3}$  & $0.20$ & $0.20$ & $0.20$ & $0.20$ & $0.20$ \\
D$7$ &  $10^{-3}$& $10^{-3}$ & $10^{-3}$ & $10^{-3}$  & $10^{-3}$ &  $10^{-3}$ & $0.25$ & $0.25$ & $0.25$ & $0.25$   \\
D$8$ &  $10^{-3}$ & $10^{-3}$ &  $10^{-3}$ & $10^{-3}$  & $10^{-3}$ &  $10^{-3}$ & $10^{-3}$ & $0.33$  & $0.33$ & $0.33$  \\
D$9$ &  $10^{-3}$ & $10^{-3}$ & $10^{-3}$ & $10^{-3}$  & $10^{-3}$ &  $10^{-3}$ & $10^{-3}$ &  $10^{-3}$ & $0.49$   & $0.49$  \\
D$10$ & $10^{-3}$& $10^{-3}$ &  $10^{-3}$ & $10^{-3}$  & $10^{-3}$ & $10^{-3}$ &$10^{-3}$ &  $10^{-3}$ & $10^{-3}$   & $0.99$   \\
\bottomrule
\end{tabular}
\label{supp_tab:class_dist}
\end{table*}

\subsection{Sample quality}\label{app:samples}

Here we show some sample images generated by \acrshort{DCGAN} and Pres\acrshort{GAN}, together with real images from each dataset. These images were not cherry-picked, we randomly selected samples from all models. For Pres\acrshort{GAN}, we show the mean of the generator distribution, conditioned on the latent variable $z$.
In general, we observed the best image quality is achieved by the entropy-regularized Pres\acrshort{GAN}.

\newpage

\begin{figure}[t]
	\centering
	\subfigure[Real images.]{\includegraphics[width=\textwidth, height=2.5cm]{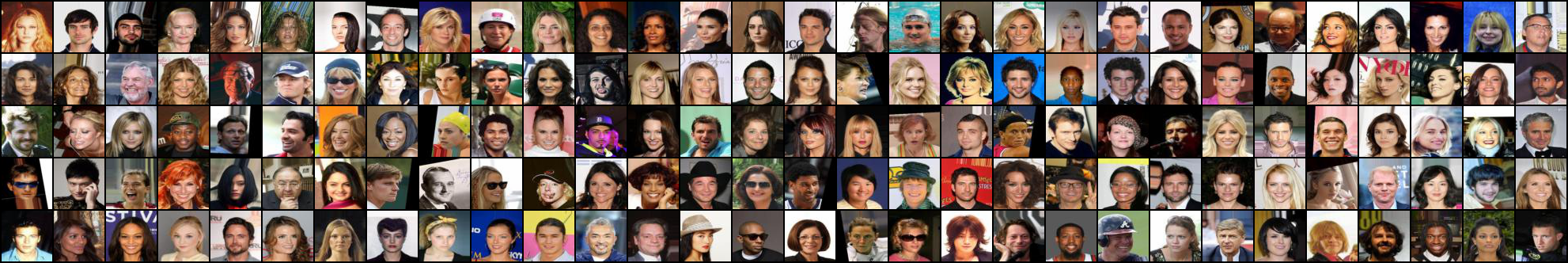}}
	\subfigure[\acrshort{DCGAN} Samples]{\includegraphics[width=\textwidth, height=2.5cm]{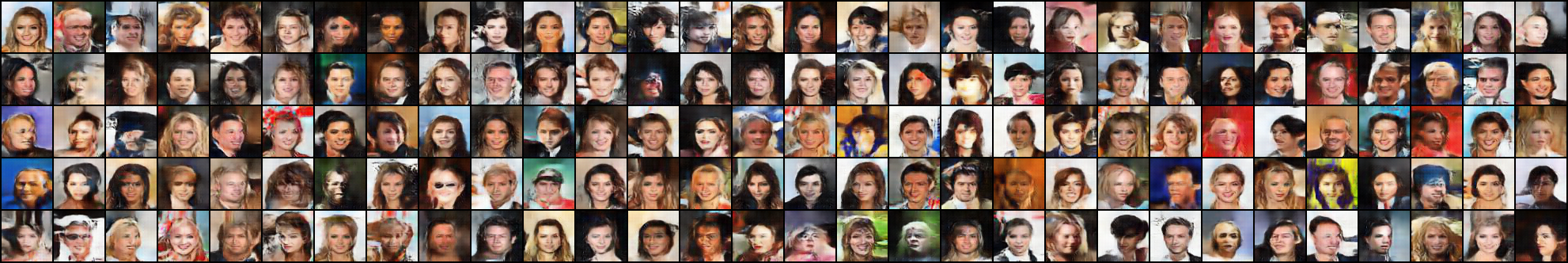}}
	\subfigure[\acrshort{VEEGAN} Samples]{\includegraphics[width=\textwidth, height=2.5cm]{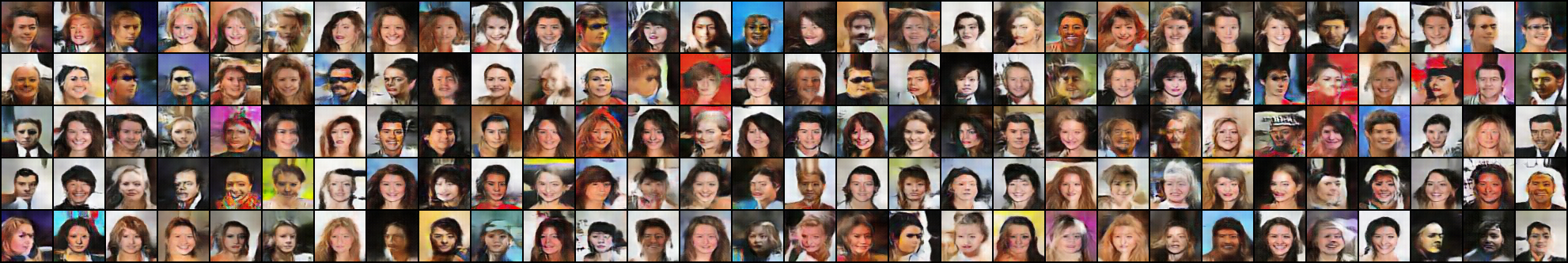}}
	\subfigure[\acrshort{PACGAN} Samples]{\includegraphics[width=\textwidth, height=2.5cm]{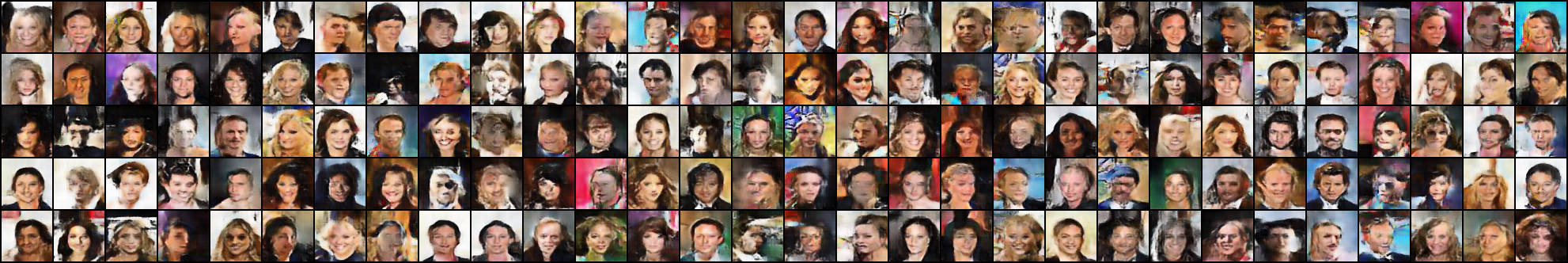}}
	\subfigure[Pres\acrshort{GAN} Samples]{\includegraphics[width=\textwidth, height=2.5cm]{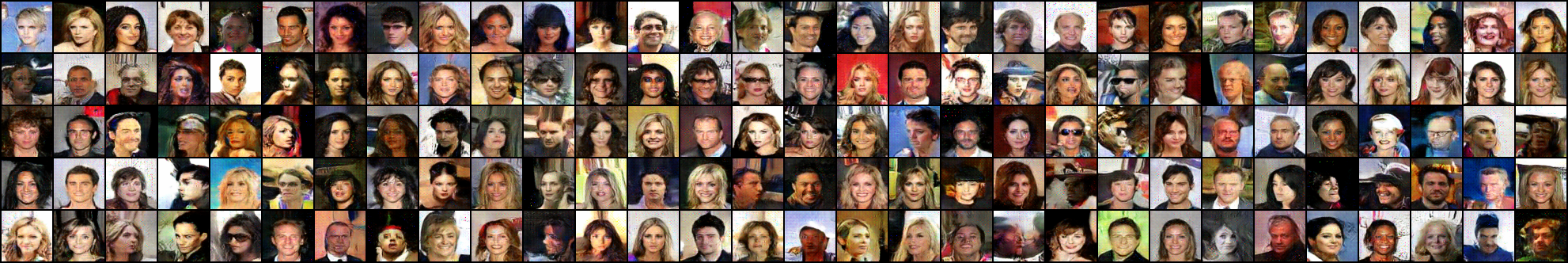}}
	\caption{Real and generated images on \textsc{celeba}.}
	\label{supp_fig:images_celeba}
\end{figure}

\end{document}